\documentclass[runningheads,table]{llncs}

\usepackage{etoolbox}
\newbool{archiveversion}
\booltrue{archiveversion}
\newbool{draftversion}

\ifbool{draftversion}{\usepackage{todonotes}}{\usepackage[disable]{todonotes}}
\usepackage{amsmath,amssymb,amsfonts,xcolor,graphicx,wrapfig,paralist}
\usepackage{algorithmicx,algorithm}
\usepackage[noend]{algpseudocode}
\usepackage{mathtools}
\usepackage{booktabs}
\usepackage{adjustbox}
\usepackage{xspace}
\usepackage{enumitem}
\usepackage{standalone}
\usepackage{tikz-uml}
\usetikzlibrary{arrows}
\usetikzlibrary{arrows.meta}
\usepackage{pgfplots}
\usepackage{array}
\usepackage{csvsimple}
\usepackage{multirow}
\usepackage[pdftex]{hyperref}
\usepackage{xcolor}
\usepackage{siunitx-v2}

\definecolor[named]{ACMBlue}{cmyk}{1,0.1,0,0.1}
\definecolor[named]{ACMYellow}{cmyk}{0,0.16,1,0}
\definecolor[named]{ACMOrange}{cmyk}{0,0.42,1,0.01}
\definecolor[named]{ACMRed}{cmyk}{0,0.90,0.86,0}
\definecolor[named]{ACMLightBlue}{cmyk}{0.49,0.01,0,0}
\definecolor[named]{ACMGreen}{cmyk}{0.20,0,1,0.19}
\definecolor[named]{ACMPurple}{cmyk}{0.55,1,0,0.15}
\definecolor[named]{ACMDarkBlue}{cmyk}{1,0.58,0,0.21}

\hypersetup{
	colorlinks,
	linkcolor=ACMPurple,
	citecolor=ACMPurple,
	urlcolor=ACMDarkBlue,
	filecolor=ACMDarkBlue}

\usepackage{breakcites} 

\DeclareFixedFont{\ttb}{T1}{txtt}{bx}{n}{8} 
\DeclareFixedFont{\ttm}{T1}{txtt}{m}{n}{8}  
\DeclareFixedFont{\tti}{T1}{txtt}{it}{n}{8}

\usepackage{color}
\definecolor{deepblue}{rgb}{0,0,0.5}
\definecolor{deepred}{rgb}{0.6,0,0}
\definecolor{deepgreen}{rgb}{0,0.5,0}

\usepackage{listings}

\usepackage[capitalise]{cleveref}

\usepackage[compatibility=false]{caption}
\usepackage{subcaption}
\usepackage{placeins} 

\usepackage{booktabs, multirow} 

\usepackage[T1]{fontenc}

\usepackage[numbers,sort]{natbib} 
\usetikzlibrary{shapes,positioning,arrows,calc,automata,matrix,fit}
\tikzstyle{state}+=[minimum size = 6mm, inner sep=0,outer sep=1]
\tikzset{->,>=stealth'}

\pgfdeclarelayer{nodelayer}
\pgfdeclarelayer{edgelayer}
\pgfsetlayers{edgelayer,nodelayer,main,connections,background}

\input{format/macros}

\tikzstyle{new style 0}=[fill=none, draw=none, shape=circle]
\tikzstyle{none}=[fill=none, draw=none, shape=circle]
\tikzstyle{left-align}=[fill=none, shape=circle, align=left]
\tikzstyle{center-align}=[fill=none, shape=circle, align=center]
\tikzstyle{rectangle}=[fill={rgb,255: red,232; green,232; blue,232}, draw=black, shape=rectangle, minimum width=4cm, minimum height=1cm]
\tikzstyle{rectangle-large}=[fill={rgb,255: red,232; green,232; blue,232}, draw=black, shape=rectangle, minimum width=4.8cm, minimum height=1cm]
\tikzstyle{center-align}=[fill=none, draw=none, shape=circle, align=center]
\tikzstyle{rectangle-white}=[fill=white, draw=black, shape=rectangle, align=center, minimum width=3cm, minimum height=1.5cm]
\tikzstyle{dot-line}=[dotted]
\tikzstyle{white}=[fill=white]

\tikzstyle{white-box}=[fill=white, -, loosely dashed]
\tikzstyle{arrow}=[->]
\tikzstyle{line}=[-, fill=white]
\tikzstyle{green-box-line}=[-, fill={rgb,255: red,197; green,245; blue,186}, opacity=0.2]
\tikzstyle{thin line}=[-, line width=1pt]
\tikzstyle{yellow-box-line}=[-, fill={rgb,255: red,247; green,255; blue,170}]
\tikzstyle{arrow 2}=[->]
\tikzstyle{new edge style 0}=[->]
\tikzstyle{blue box}=[-, fill={rgb,255: red,208; green,241; blue,245}]
\tikzstyle{gray-box}=[-, fill={rgb,255: red,244; green,244; blue,244}]
\tikzstyle{arrow-uthick}=[->, ultra thick]
\tikzstyle{arrow-vthick}=[->, very thick]
\tikzstyle{dot-edge}=[dotted, {|-|}]
\tikzstyle{dot-white}=[-, dotted, fill=white]

\title{Monitizer: Automating Design and Evaluation of Neural Network Monitors\thanks{This research was funded in part by the German Research Foundation (DFG) project
		{427755713 GOPro} and the MUNI Award in Science and Humanities 
		{MUNI/I/1757/2021} of the Grant Agency of Masaryk University.}}
\author{Muqsit Azeem\inst{1}\,\orcidID{0000-0003-4532-8344} \and
	Marta Grobelna\inst{1}\,\orcidID{0009-0003-3314-3358} \and
	Sudeep Kanav\inst{2}\,\orcidID{1111-2222-3333-4444} \and
	Jan K\v{r}et\'{i}nsk\'{y}\inst{2,1}\,\orcidID{0000-0002-8122-2881} \and \\
	Stefanie Mohr\inst{1}\,\orcidID{0000-0002-8630-3218} \and
	Sabine Rieder\inst{1,2,3}\,\orcidID{0009-0006-6397-3100}}
	
\institute{Technical University of Munich, Munich, Germany \and
	Masaryk University, Brno, Czech \and Audi AG, Ingolstadt, Germany}

\begin{document}

\maketitle

\begin{abstract}
	The behavior of neural networks (NNs) on previously unseen types of data (out-of-distribution or OOD) is typically unpredictable. 
	This can be dangerous if the network's output is used for decision making in a safety-critical system.
	Hence, detecting that an input is OOD is crucial for the safe application of the NN.
	Verification approaches do not scale to practical NNs, making runtime monitoring more appealing for practical use.	
	While various monitors have been suggested recently, their optimization	for a given problem, as well as comparison with each other and reproduction of results, remain challenging.
	
	We present a tool for users and developers of NN monitors.
	It allows for  
	(i)~application of various types of monitors from the literature to a given input NN, 
	(ii)~optimization of the monitor's hyperparameters, and 
	(iii)~experimental evaluation and comparison to other approaches.
	Besides, it facilitates the development of new monitoring approaches.
	We demonstrate the tool's usability on several use cases of different types of users as well as on a case study comparing different approaches from recent literature.
\end{abstract} 

\section{Introduction}
\paragraph{Neural networks (NNs)} are increasingly used in safety-critical applications due to their good performance even on complex problems.
However, their notorious unreliability makes their safety assurance even more important.
In particular, even if the NN is well trained on the data that it is given and works well on similar data (so-called \emph{in-distribution (ID) data}), it is unclear what it does if presented with a significantly different input (so-called \emph{out-of-distribution (OOD) data}).
For instance, what if an NN for traffic signs recognition trained on pictures taken in Nevada is now presented with a traffic sign in rainy weather, a European one, or a billboard with an elephant?

To ensure safety in all situations, we must at least recognize that the input is OOD; thus, the network's answer is unreliable, no matter its confidence.
Verification, a classic approach for proving safety, is extremely costly and essentially infeasible for practical NNs \cite{VnnComp3}.
Moreover, it is mainly done for ID or related data \cite{robustness,VnnComp3}.
For instance, robustness is typically proven for neighborhoods of essential points, which may ensure correct behavior in the presence of noise or rain, but not elephants~\cite{Henriksen2020,katz2022reluplex,marabou,Mueller2022}.
In contrast, runtime verification and particularly runtime \emph{monitoring} provide a cheap alternative.
Moreover, the industry also finds it appealing as it is currently the only formal-methods approach applicable to industrial-sized NNs.

\paragraph{OOD runtime monitoring methods} have recently started flourishing ~\cite{henzinger2020outside,gaussObj,Cheng2019,liu2020energy,sun2022dice,hsu2020generalized}.
Such a runtime monitor tries to detect if the current input to the NN is OOD.
To this end, it typically monitors the behavior of the network (e.g., the output probabilities or the activation values of the neurons) and evaluates whether 
the obtained values resemble the ones observed on known ID data.
If not, the monitor raises an alarm to convey suspicion about OOD data.
\vspace{-3mm}
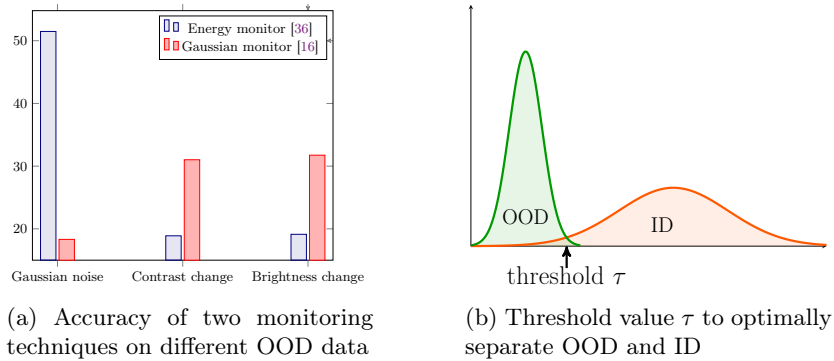
\begin{figure}
	\centering
	\begin{subfigure}{0.4\textwidth}
		\centering
		\resizebox{\textwidth}{!}{\pgfplotstableread[row sep=\\,col sep=&]{
    OOD data & Energy & Gaussian \\
    Gaussian noise     & 51.48 & 18.32 \\
    Contrast change     & 18.89 & 31.02  \\
    Brightness change  & 19.13 & 31.75 \\
    }\tabledata 

\begin{tikzpicture}
  \begin{axis}[
            ybar,
            symbolic x coords={Gaussian noise,Contrast change,Brightness change},
            xtick=data
        ]
        \addplot[blue!50!black, fill=blue!50!black, fill opacity=0.1] table[x=OOD data,y=Energy]{\tabledata};
        \addplot table[x=OOD data,y=Gaussian]{\tabledata};
        \legend{Energy monitor \cite{liu2020energy}, Gaussian monitor \cite{gaussMon}}
    \end{axis}
\end{tikzpicture}}
		\caption{Accuracy of two monitoring techniques on different OOD data}
		\label{fig:monitor-comp-challenge}
	\end{subfigure}
	\hspace{10mm}	
	\begin{subfigure}{0.4\textwidth}
		\centering
		\resizebox{\textwidth}{!}{\pgfmathdeclarefunction{gauss}{2}{%
	\pgfmathparse{1/(#2*sqrt(2*pi))*exp(-((x-#1)^2)/(2*#2^2))}%
}

\begin{tikzpicture}
	\def\meana{0}
	\def\meanb{2.7}
	\def\stda{0.3}
	\def\stdb{1}
	\def\xmin{-1}
	\def\xmax{5.5}
	
	\begin{axis}[
		width=10cm,
		height=7.3cm,
		clip=false,
		font=\Large,
		domain=\xmin:\xmax, 
		samples=50,
		every axis plot post/.append style={-,mark=none,smooth}, 
		xtick=\empty,
		ytick=\empty,
		axis x line=bottom, 
		axis y line=left, 
		ymax=1.65,
		x tick label style={major tick length=0pt}
		] 
		\addplot [draw=none, fill=green!60!black, opacity=0.1] {gauss(\meana, \stda)} \closedcycle;
		\addplot [draw=none, fill=orange!70!red, opacity=0.1] {gauss(\meanb, \stdb)} \closedcycle;
		\addplot[color=orange!70!red,line width=1.5pt] {gauss(\meanb,\stdb)};
		\addplot[color=green!60!black,line width=1.5pt,domain=-1:1] {gauss(\meana,\stda)};
		\node (a) at (100,20) {OOD};
		\node (b) at (350,15) {ID};
		\node ()  at (175,-20) {\LARGE{threshold $\tau$}};
		\draw[->,line width=1.5pt](175,-15)--(175,0);
	\end{axis}
\end{tikzpicture}}
		\caption{Threshold value $\tau$ to optimally separate OOD and ID}
		\label{fig:energy-threshold-intro}
	\end{subfigure}	
	\caption{Illustration of challenges for OOD detection}
\end{figure}
\vspace{-5mm}

\paragraph{Challenges:} While this approach has demonstrated potential, several practical issues arise:
\begin{itemize}[leftmargin=*]
	\item How can we \emph{compare} two monitors and determine which one is better? 
	Considering the example of autonomous driving, an OOD input could arise from the fact that some noise was introduced by
	sensors or the brightness of the environment was perturbed. 
	A monitor might perform well on one kind of OOD input but may not on another~\cite{Tajwar2021}, as 
	better performance in one class of OOD data does not imply the same in another class (see \cref{fig:monitor-comp-challenge}).

	\item Applying a particular monitoring technology to a concrete NN involves significant tweaking and \emph{hyperparameter tuning}, with no push-button technology available.
	OOD monitors typically compute a value from the input and the behavior of the NN.
	The input is considered OOD if this value is smaller than a configurable \emph{threshold} $\tau$ (see \cref{fig:energy-threshold-intro}).
	The value of this threshold has a significant influence on the performance of the monitors.
	More inputs would be classified as OOD if the threshold value is high, and vice versa.
	Moreover, OOD monitors generally have \emph{multiple parameters} that require tuning, thereby aggravating the complexity of manual configuration.

	\item As OOD monitoring can currently be described as a search for a good heuristic, many more heuristics will appear, implying the need for streamlining their handling and fair comparison. 
\end{itemize}
In this paper, we provide the infrastructure for users and developers of NN monitors aiming at detecting OOD inputs (onwards just ``monitors'').

\paragraph{Our contributions} can be summarized as follows:
\begin{itemize}[leftmargin=*]
	\item We provide a modular tool called \monitizer for automatic learning/constructing, optimizing, and evaluating monitors.
	\item \monitizer supports (i) \emph{easy practical use}, providing various recent monitors from the literature, which can directly be optimized and applied to user-given networks and datasets
	with no further inputs required; the push-button solution offers automatic choice of the best available monitor without requiring any knowledge on the side of the user;
	(ii) \emph{advanced development use}, with the possibility of easily integrating a new monitor or new evaluation techniques.
	The framework also foresees and allows for the integration of monitoring other properties than OOD.
	\item We provide a library of 19 well-known monitors from the scientific literature to be used off-the-shelf, accompanied by 9 datasets and 15 NNs, which can be used for easy but rich automatic evaluation and comparison of monitors on various OOD categories.
	\item We demonstrate the functionality for principled use cases accompanied by examples and a case study comparing a few recent monitoring approaches.
\end{itemize}
Altogether, we are giving users the infrastructure for automatic creation of monitors,
development of new methods, and their comparison to similar approaches.

\section{Related Work}
\label{sec:rw}
\inlineheadingbf{NN monitoring frameworks}
\openood~\cite{yang2022openood,zhang2023openood} contains task-specific benchmarks for OOD detection
that consist of an ID and multiple OOD datasets for specific tasks (e.g., Open Set Recognition and Anomaly Detection).
Both \openood and \monitizer contain several different monitors
and benchmarks.
\monitizer provides functionality to tune the monitors for the given objective,
supports a comprehensive evaluation of monitors on a specific ID dataset
by automatically providing generated OOD inputs by, e.g., the addition of noise, 
and can easily be extended with more datasets.
\openood, in contrast to \monitizer, does not support hyperparameter tuning and generation of OOD inputs.

Samuels et al. propose a framework to optimize an OOD monitor during runtime on newly experienced OOD inputs~\cite{pmlr-v162-katz-samuels22a}.
While this contains optimization, the framework is specific to one monitor
and is based on active learning.
\monitizer is meant to work in an offline setting and optimize a monitor before it is deployed. Additionally, \monitizer is built for extensibility and reusability, which the other tool is not, e.g., it lacks an executable.

\pytorchood~\cite{PyTorch-OOD} is a library for OOD detection, yet despite its name, it is \emph{not} part of the official PyTorch-library.
It includes several monitors, datasets,
and supports the evaluation of the integrated monitors.
Both \monitizer and \pytorchood provide a library of monitors and datasets. 
However, there are significant differences.
\monitizer supports optimization of monitors, allowing us to return monitors optimal for a chosen objective,
provides a more structured view of the dataset,
and provides a transparent and detailed evaluation showing how a monitor performs on different OOD classes.
Besides, we provide a one-click solution to easily evaluate the whole set of monitors
and automatically return the best available option, fine-tuned to the case.
Consequently, \monitizer is a tool that is much easier to use and extend.
Last but not least, it is an alternative implementation that allows cross-checking outcomes,
thereby making monitoring more trustworthy.

\inlineheadingbf{OOD benchmarking}
Various datasets have been published for OOD benchmarking~\cite{hendrycks2022scaling,LostAndFound,hendrycks2019benchmarking,Olber2023,Henriksson2019},
Breitenstein et al. present a classification for different types
of OOD data in automated driving~\cite{Breitenstein2020},
and Ferreira et al. propose a benchmark set for OOD with several different categories~\cite{Ferreira2021}.

\section{Monitizer}
\label{sec:implementation}
\monitizer aims to assist the developers and users of NN monitors and developers of new monitoring techniques
by supporting optimization and transparent evaluation of their monitors.
It structures OOD data in a hierarchy of classes, and a monitor can be tuned for any (combination) of these classes.
It also provides a one-click solution to evaluate a set of monitors and return the best available option optimized for the given requirement.

\subsection{Overview}
\begin{figure}[t]
	\makebox[\textwidth][c]{\resizebox{1\textwidth}{!}{\input{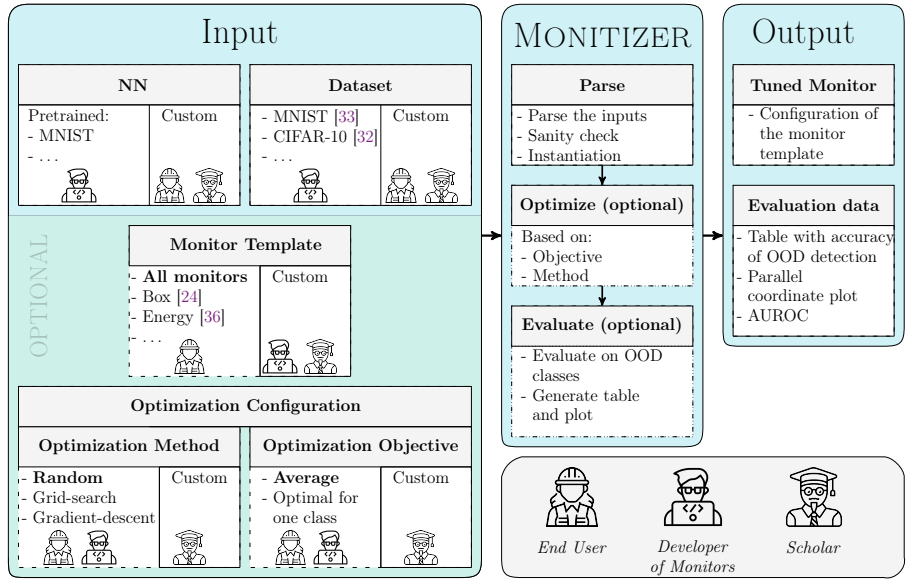}}}
	\caption{Architecture of \monitizer:
	The required inputs are an NN and the dataset (both can be chosen from existing options).
	The dashed area indicates optional inputs, and the bold-faced option indicates the default value.
	The icons\protect\footnotemark{} indicate which types of users are expected to use each of the options.}
	\vspace*{-4mm}
	\label{fig:overview-monitizer}
\end{figure}

\monitizer offers two main building blocks, as demonstrated in \cref{fig:overview-monitizer}: optimization and evaluation of NN monitors.
NN monitors are typically parameterized and usually depend on the NN and dataset.
Before one can evaluate them, they need to be configured and possibly tuned.
We refer to monitors that are not yet configured as \emph{monitor templates}.
\monitizer optimizes the monitor templates and evaluates them afterward on several different OOD classes, i.e., types of OOD data.

\monitizer needs at least two inputs (see \cref{fig:overview-monitizer}): an NN, and an ID-dataset.
The user can also provide a monitor template and an optimization configuration (consisting of an optimization objective and optimization method).
If these are not provided, \monitizer reverts to the default values (i.e., evaluating all monitors using the AUROC-score without optimization).
For both inputs, the user can choose from the options we offer or provide a custom implementation.

\monitizer optimizes the provided monitor based on the optimization objectives and method on the given ID dataset.
An example of optimization would be:
\footnotetext{Thanks to Flaticon.com for the Icons}
maximize the detection accuracy on blurry images, but keep the accuracy on ID images at least \SI{70}{\percent}.
Optimization is necessary to obtain a monitor that is ready to use.
However, it is possible to evaluate a monitor template on its default values for the parameters 
using the \emph{AUROC}-score (Area Under the Receiver Operating Characteristic Curve)\footnote{The ROC (Receiver Operating Characteristic) curve shows the performance of a binary classifier with different decision thresholds. The AUROC computes the area under this curve. The best possible value is 1, indicating perfect prediction.}.

On successful execution, 
\monitizer provides the user with a configuration of the monitor template 
and the evaluation result.
This can be either a table with the accuracy of OOD detection for each OOD dataset along with a parallel coordinate plot for the same (in case of optimization) or the AUROC score. 

\subsection{Use Cases}\label{sec:use-cases}
We envision three different types of users for \monitizer:
\begin{enumerate}
	\item \textbf{The End User}
	
	\textit{Context:}  The end user of a monitor,
	e.g., an engineer in the aviation industry,
	is interested in the end product,
	not in the intricacies of the underlying monitoring technique.
	She intends to evaluate one or all monitors provided by \monitizer
	for her custom NN and dataset, and 	wants to 
	come to a conclusion on which one to use.
	She has an NN that needs to be monitored. 
	Additionally, she has her own proprietary ID dataset,
	e.g., the one on which the NN was trained.
	She wants a monitor fulfilling some requirement, e.g., one that is optimal on average for all classes or one that can detect a specific type of OOD that her NN is not able to handle properly. 

	\textit{Usage:} 
	Such a user can obtain a monitor tuned to her needs using \monitizer without much effort.
	\monitizer supports this feature out of the box.
	It provides various monitors (19 at present) that can be optimized for a given network. 
	In case she wants to use a custom NN or a dataset, she has to provide the NN as PyTorch-dump or in onnx-format~\cite{onnx}
	and add 
	some lines of code
	to implement the interface for loading her data.
	
	\textit{Required Effort:} 
	After providing the interface for her custom dataset, the user only has to trigger the execution.
	The execution time depends on the hardware quality, the NN's size, the chosen monitor's complexity, and the dataset's size.
	
	\item \textbf{The Developer of Monitors}
	
	\textit{Context:} The developer of monitoring techniques, e.g., 
	a researcher working in runtime verification of NNs,
	aims to create novel techniques and 
	assess their performance in comparison to established methods. 
	
	\textit{Usage:} Such a user can plug their novel monitor into \monitizer and evaluate it.
	\monitizer directly provides the most commonly used NNs and datasets for academic evaluation.
	
	\textit{Required Effort:} The code for the monitor needs to be in Python and should 
	implement the functions specified in the interface for monitors in \monitizer. 
	Afterward, she can trigger the evaluation of her monitoring technique.
	
	\item \textbf{The Scholar}
	
	\textit{Context:}
	An expert in monitoring, 
	e.g., an experienced researcher in NN runtime verification,
	intends to explore beyond the current boundaries.
	She might want to adapt an NN monitor to properties other than OOD,
	or to experiment with custom NNs or datasets.
	
	\textit{Usage:} \monitizer provides interfaces, and instructions
	on how to integrate new NNs, datasets, monitors,
	custom optimization methods and objectives.
	
	\textit{Required Effort:} The required integration effort depends on the complexity of the concrete use case.
	For example, adding an NN would take much less time than developing a new monitor.
\end{enumerate}
More detailed examples are available in \myrefappendix{app:example-use-cases}.

\subsection{Phases of \monitizer}
An execution of \monitizer is typically a sequence of three phases: 
parse, optimize, and evaluate. As mentioned, the user can decide to skip the optimization or the evaluation.

\inlineheadingbf{Parse}
This phase parses the input, loads the NN and dataset, and instantiates the monitor.
It also performs sanity checks on the inputs, e.g.,
the datasets are available in the file system, the provided monitor is implemented correctly, etc.

\inlineheadingbf{Optimize}\label{sec:optimization}
This phase tunes the parameters of a given monitor template to maximize an objective.
It depends on two inputs, the optimization method and the optimization objective, that the user has to give.

An illustrative depiction of this process can be found in \myrefappendix{app:optimization}.
The optimization method defines the search space and generates a new candidate monitor by setting its parameters. 
\monitizer then uses the optimization objective to evaluate this candidate.
If the objective is to optimize at least one OOD class, \monitizer evaluates the monitor on a validation set of this class, which is distinct from the test set used in the evaluation later.
The optimization method obtains this result and decides whether to continue optimizing or stop and return the best monitor that it has found.

\monitizer provides three optimization methods: random, grid-search, and gradient descent.
Random search tries out a specified number of random sets of parameters and returns the monitor that worked best among these.
Grid-search specifies a search grid by looking at the minimal and maximal values of the parameters. 
It then defines a grid on the search space.
The monitor is infused with these parameters for each grid vertex and evaluated on the objective.
Gradient-descent follows the gradient of the objective function towards the optimum. 

\monitizer supports multi-objective optimization of monitors.
A user can specify a set of OOD classes to optimize for and the minimum required accuracy for ID detection.
Single objective optimization is a special case when only one OOD class is specified for optimization.
Based on a configuration value,
\monitizer would generate a set of different weight combinations for the objectives
and
create and evaluate a monitor for each of these combinations.
If there are two objectives, \monitizer generates a Pareto frontier plot;
in the case of more than two objectives, the tool generates a table. 
The user obtains the performance of the optimized monitor for each weight-combination of objectives. 

\inlineheadingbf{Evaluate}
\begin{figure}[t!]
	\centering
	\resizebox{\textwidth}{!}{\begin{tikzpicture}[{uml-arrow/.style}={->, -{Triangle[open,scale=2]}}]
	\begin{pgfonlayer}{nodelayer}
		\node [style=none] (4) at (-1.25, 7.25) {};
		\node [style=rectangle, minimum width=1.0cm, minimum height=1.2cm, align=center, text width=2cm, font={\large}, fill=none] (5) at (-1.25, 8) {OOD Class};
		\node [style=rectangle, minimum width=1.0cm, minimum height=1.2cm, align=center, text width=2cm, font={\large}, fill=none] (34) at (-7.25, 6) {Generated OOD};
		\node [style=rectangle, minimum width=1.0cm, minimum height=1.2cm, align=center, text width=2cm, font={\large}, fill=none] (42) at (4.5, 6) {Collected OOD};
		\node [style=rectangle, minimum width=1.0cm, minimum height=1.2cm, align=center, text width=2.4cm, font={\large}, fill=none] (11) at (-11.25, 4) {Perturbation};
		\node [style=rectangle, minimum width=1.0cm, minimum height=1.2cm, align=center, text width=2cm, font={\large}, fill=none] (29) at (-5.125, 4) {Noise};
		\node [style=rectangle, minimum width=1.0cm, minimum height=1.2cm, align=center, text width=2.2cm, font={\large}, fill=none] (47) at (-1.25, 4) {Wrong Prediction};
		\node [style=rectangle, minimum width=1.0cm, minimum height=1.2cm, align=center, text width=1.8cm, font={\large}, fill=none] (17) at (-13.75, 2) {Light};
		\node [style=rectangle, minimum width=1.0cm, minimum height=1.2cm, align=center, text width=1.8cm, font={\large}, fill=none] (23) at (-12.5, 0) {Contrast};
		\node [style=rectangle, minimum width=1.0cm, minimum height=1.2cm, align=center, text width=1.8cm, font={\large}, fill=none] (50) at (-11.25, 2) {Gaussian Blur};
		\node [style=rectangle, minimum width=1.0cm, minimum height=1.2cm, align=center, text width=1.8cm, font={\large}, fill=none] (62) at (-10, 0) {Color Inversion};
		\node [style=rectangle, minimum width=1.0cm, minimum height=1.2cm, align=center, text width=1.8cm, font={\large}, fill=none] (69) at (-8.75, 2) {Rotation};
		\node [style=rectangle, minimum width=1.0cm, minimum height=1.2cm, align=center, text width=1.8cm, font={\large}, fill=none] (35) at (-6.25, 2) {Gaussian};
		\node [style=rectangle, minimum width=1.0cm, minimum height=1.2cm, align=center, text width=1.8cm, font={\large}, fill=none] (41) at (-4, 2) {Salt and Pepper};
		\node [style=rectangle, minimum width=1.0cm, minimum height=1.2cm, align=center, text width=2.2cm, font={\large}, fill=none] (71) at (-1.25, 2) {Adversarial Example};
		\node [style=rectangle, minimum width=1.0cm, minimum height=1.2cm, align=center, text width=2.2cm, font={\large}, fill=none] (65) at (-1.25, 0) {FGSM};
		\node [style=rectangle, minimum width=1.0cm, minimum height=1.2cm, align=center, text width=2.2cm, font={\large}, fill=none] (77) at (1.75, 4) {Unseen Object};
		\node [style=rectangle, minimum width=1.0cm, minimum height=1.2cm, align=center, text width=2.2cm, font={\large}, fill=none] (83) at (4.5, 4) {New Environment};
		\node [style=rectangle, minimum width=1.0cm, minimum height=1.2cm, align=center, text width=2.2cm, font={\large}, fill=none] (95) at (7.25, 4) {New World};
		\node [style=rectangle, minimum width=1.0cm, minimum height=1.2cm, align=center, text width=2.2cm, font={\large}, fill=none] (89) at (4.5, 2) {Weather};
		\node [style=none] (188) at (4.5, 7) {};
		\node [style=none] (133) at (-7.25, 7) {};
		\node [style=none] (170) at (1.75, 5) {};
		\node [style=none] (171) at (4.5, 5) {};
		\node [style=none] (173) at (7.25, 5) {};
		\node [style=none] (174) at (-7.25, 5) {};
		\node [style=none] (175) at (-1.25, 5) {};
		\node [style=none] (176) at (-11.25, 5) {};
		\node [style=none] (177) at (-1.25, 7) {};
		\node [style=none] (178) at (-5.125, 3) {};
		\node [style=none] (179) at (-6.25, 3) {};
		\node [style=none] (180) at (-4, 3) {};
		\node [style=none] (181) at (-12.5, 3) {};
		\node [style=none] (182) at (-11.25, 3) {};
		\node [style=none] (183) at (-8.75, 3) {};
		\node [style=none] (184) at (-10, 3) {};
		\node [style=none] (186) at (-13.75, 3) {};
		\node [style=none] (187) at (-5.125, 5) {};
	\end{pgfonlayer}
	\begin{pgfonlayer}{edgelayer}
		\draw [style=uml-arrow] (177.center) to (5);
		\draw [style=line] (188.center) to (177.center);
		\draw [style=line] (177.center) to (133.center);
		\draw [style=line] (133.center) to (34);
		\draw [style=line] (188.center) to (42);
		\draw [style=uml-arrow] (171.center) to (42);
		\draw [style=line] (171.center) to (170.center);
		\draw [style=line] (170.center) to (77);
		\draw [style=line] (171.center) to (83);
		\draw [style=line] (171.center) to (173.center);
		\draw [style=line] (173.center) to (95);
		\draw [style=uml-arrow] (89) to (83);
		\draw [style=uml-arrow] (174.center) to (34);
		\draw [style=line] (174.center) to (176.center);
		\draw [style=line] (174.center) to (175.center);
		\draw [style=uml-arrow] (178.center) to (29);
		\draw [style=line] (178.center) to (179.center);
		\draw [style=line] (178.center) to (180.center);
		\draw [style=line] (175.center) to (47);
		\draw [style=uml-arrow] (71) to (47);
		\draw [style=uml-arrow] (65) to (71);
		\draw [style=line] (176.center) to (11);
		\draw [style=uml-arrow] (182.center) to (11);
		\draw [style=line] (182.center) to (184.center);
		\draw [style=line] (184.center) to (183.center);
		\draw [style=line] (181.center) to (182.center);
		\draw [style=line] (186.center) to (181.center);
		\draw [style=line] (186.center) to (17);
		\draw [style=line] (181.center) to (23);
		\draw [style=line] (182.center) to (50);
		\draw [style=line] (184.center) to (62);
		\draw [style=line] (183.center) to (69);
		\draw [style=line] (179.center) to (35);
		\draw [style=line] (180.center) to (41);
		\draw [style=line] (187.center) to (29);
	\end{pgfonlayer}
\end{tikzpicture}}
	\caption{Class diagram depicting the different types of OOD data.
	}
	\label{fig:classes}
\end{figure}
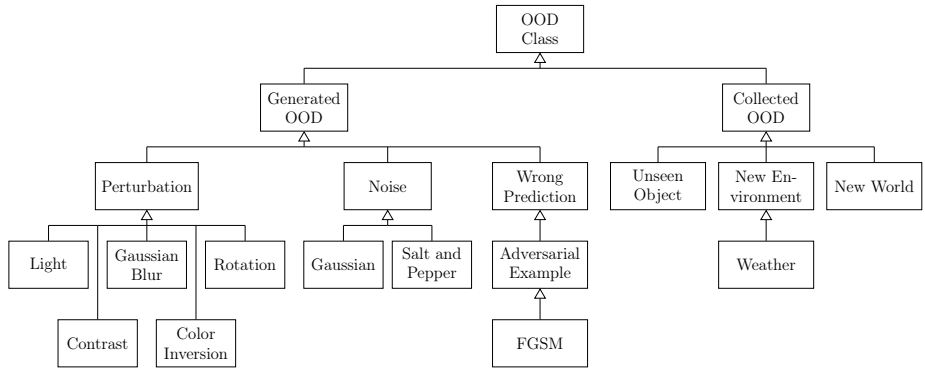
The evaluation of NN monitors in \monitizer is structured according to the OOD classification (detailed in the next section).
We introduce this classification of OOD data 
to enable a clearer evaluation and gain knowledge about which monitor performs well on which particular class of OOD.
Typically, no monitor performs well on every class of OOD \cite{Tajwar2021}.
We highlight this in our evaluation to ensure a fair and meaningful comparison between monitors rather than restricting to a non-transparent and possibly biased average score.

After evaluation, \monitizer reports the detection accuracy for each OOD class
and can also produce a parallel-coordinates-plot displaying the reported accuracy.
\monitizer can also provide confidence intervals for the evaluation quality, which is explained in 
\myrefappendix{app:statistical-analysis}. 

\subsection{Classification of Out-of-Distribution Data}\label{sec:benchmarks}
We now introduce our classification of OOD data.
At the top level, an OOD input can either be \emph{generated}, i.e.,
obtained by distorting ID data~\cite{gaussObj,sun2021react,hendrycks2016baseline,bai2023feed,liang2018enhancing},
or it can be \emph{collected} using data from some other available dataset.
\begin{wrapfigure}[17]{l}{0.42\textwidth}
	\begin{subfigure}{0.2\textwidth}
		\centering
		\includegraphics[width=2.5cm]{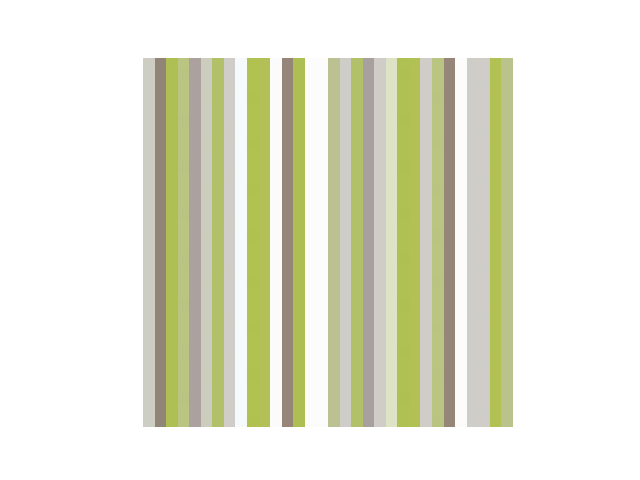}
		\caption{\scriptsize DTD~\cite{dtd}}
		\label{fig:dtd-1}
	\end{subfigure}
	\begin{subfigure}{0.2\textwidth}
		\centering
		\includegraphics[width=2.5cm]{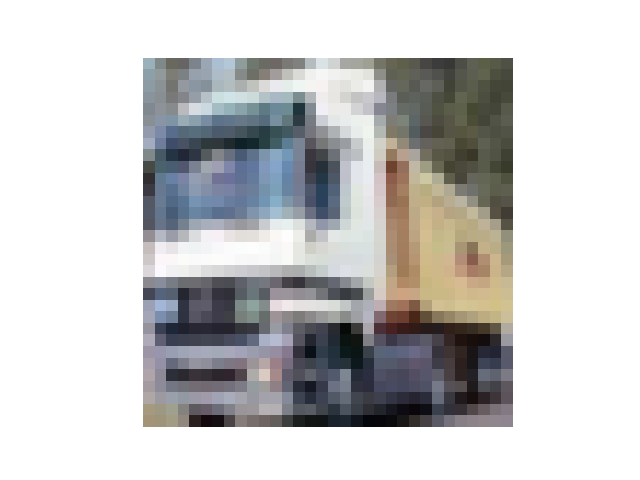}
		\caption{\scriptsize \cifarten~\cite{cifar}}
		\label{fig:cifar10-1}
	\end{subfigure}	\\
	\begin{subfigure}{0.2\textwidth}
		\centering
		\includegraphics[width=2.5cm]{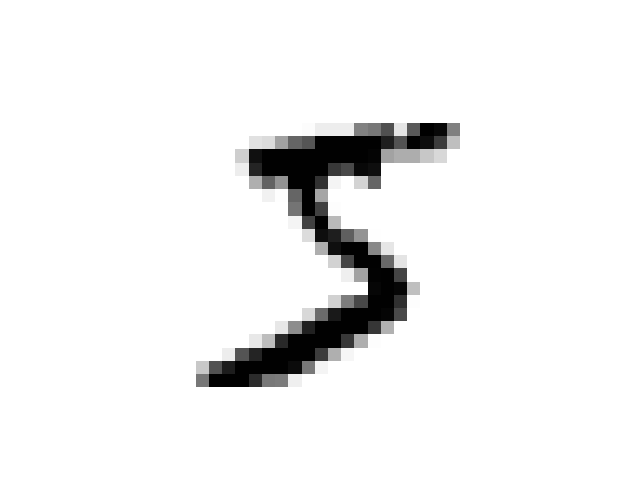}\\
		\caption{\scriptsize MNIST~\cite{mnist}}
		\label{fig:mnist-1}
	\end{subfigure}
	\begin{subfigure}{0.2\textwidth}
		\centering
		\includegraphics[width=2.5cm]{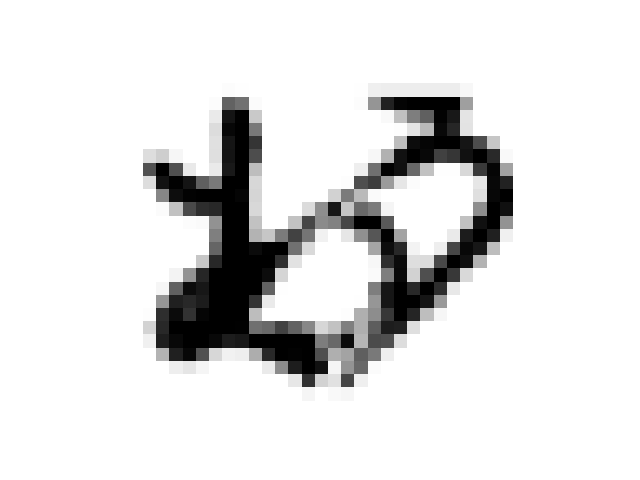}
		\caption{\scriptsize KMNIST~\cite{kmnist}}
		\label{fig:kmnist-1}
	\end{subfigure}
	\caption{Examples for OOD}
	\label{fig:OOD-example}
\end{wrapfigure}

The notion of generated OOD is straightforward. 
These classes are created by slightly distorting ID data, for example, by increasing the contrast or adding noise.
An important factor is the amount of distortion, e.g., the amount of noise, as it influences the NN's performance and needs to be high enough to transform an ID into an OOD input.

We explain the idea of collected OOD with the help of an example
shown in \cref{fig:OOD-example}. 
Consider an ID dataset that consists of textures (\cref{fig:dtd-1}).
Images containing objects (\cref{fig:cifar10-1}) differ from images showing just a texture.
But, when we consider a dataset of numbers as ID (\cref{fig:mnist-1}),
it seems much more similar to a dataset of letters (\cref{fig:kmnist-1}) 
than textures are to objects.
In the first case, the datasets have no common meaning or concept, as if they were belonging to a \emph{new world}.
In the second case, the environment and the underlying concept are similar, but an \emph{unseen object} is placed in it.

\cref{fig:classes} shows our classification of the OOD data. 
It is based on the kind of OOD data we found in the literature (discussed in \cref{sec:rw}).
\myrefappendix{app:description-of-ood} contains a detailed description of each class and an illustrative figure.

\inlineheadingbf{OOD Benchmarks Implementation}
Note that the generated OOD will be automatically created by \monitizer for any given ID dataset. 
The collected OOD data 
has to be manually selected.
We provide a few preselected datasets (for example, KMNIST \cite{kmnist} as unseen objects for MNIST \cite{mnist}) in the tool.
A user can easily add more when needed.
However, for a user like the developer of monitors, MNIST and \cifarten are often sufficient to test new monitoring methodologies, as related work has shown \cite{gaussMon,henzinger2020outside}.

\subsection{Library of Monitors, NNs, and Datasets}
\monitizer currently includes 19 monitors, accompanied by 9 datasets and 15 NNs.
In the following, we give an overview of the available options.

\paragraph{Monitors}
\monitizer provides different highly cited monitors, which are also included in other tools such as {\scshape OpenOOD}/{\scshape Pytorch-OOD}.
We extended this list by adding monitors from the formal methods community (e.g., {\scshape Box} monitor, {\scshape Gaussian} monitor).
The following monitors are available in \monitizer:
{\scshape ASH-B},{\scshape ASH-P},{\scshape ASH-S}~\cite{djurisic2022extremely},
{\scshape Box-monitor}~\cite{henzinger2020outside}, {\scshape DICE}~\cite{sun2022dice},
{\scshape Energy}~\cite{liu2020energy}, {\scshape Entropy}~\cite{macedo2021entropic},
{\scshape Gaussian}~\cite{gaussMon}, {\scshape GradNorm}~\cite{huang2021importance},
{\scshape KL Matching}~\cite{hendrycks2022scaling}, {\scshape KNN}~\cite{sun2022out},
{\scshape MaxLogit}~\cite{zhang2023decoupling},
{\scshape MDS}~\cite{lee2018simple}, {\scshape Softmax}~\cite{hendrycks2016baseline},
{\scshape ODIN}~\cite{liang2018enhancing}, {\scshape ReAct}~\cite{sun2021react},
  {\scshape Mahalanobis}~\cite{ren2021simple},
{\scshape SHE}~\cite{zhang2022out}, {\scshape Temperature}~\cite{guo2017calibration}
{\scshape VIM}~\cite{wang2022vim}. 

\paragraph{Datasets} The following datasets are available in \monitizer:
CIFAR-10, CIFAR-100~\cite{cifar}, DTD~\cite{dtd}, FashionMNIST~\cite{fashionmnist}, GTSRB~\cite{gtsrb}, ImageNet~\cite{ILSVRC15}, K-MNIST~\cite{kmnist}, MNIST~\cite{mnist}, SVHN~\cite{svhn}.

\paragraph{Neural Networks}
\monitizer provides at least one pretrained NN for each available dataset.
The library contains more NNs trained on commonly used datasets in academia, such as MNIST and CIFAR-10, allowing users to evaluate monitors on different architectures. 
\myrefappendix{app:details-networks} contains a detailed description of the pretrained NNs.

\section{Summary of Evaluation by Case Study}
\label{sec:eval}

\begin{table}[t]
	\caption{Comparison of the AUROC-score of all implemented monitors on different OOD datasets multiplied by 100 (and rounded to the nearest integer). All monitors were evaluated on a fully connected network trained on MNIST.
	The cells are colored according to the relative performance of a monitor (column) in a specific OOD class (row).
	The monitors are divided in three ranks and the darker color represents better performance.
	If several monitors have the same score, they all belong to the better group.}
	\centering
	\NewDocumentCommand{\rot}{O{90} O{1em} m}{\makebox[#2][l]{\rotatebox{#1}{#3}}}%

\begin{tabular}{crrrrrrrrrrrrrrrrrrr}
	\toprule
	Perturbations & \rot{ASH-B~\cite{djurisic2022extremely}} & \rot{ASH-P~\cite{djurisic2022extremely}} & \rot{ASH-S~\cite{djurisic2022extremely}} & \rot{DICE~\cite{sun2022dice}} & \rot{Energy~\cite{liu2020energy}} & \rot{Entropy~\cite{macedo2021entropic}} & \rot{Gauss~\cite{gaussMon}} & \rot{GradNorm~\cite{huang2021importance}} & \rot{KL Matching~\cite{hendrycks2022scaling}} & \rot{KNN~\cite{sun2022out}} & \rot{MDS~\cite{lee2018simple}} & \rot{Mahalanobis~\cite{ren2021simple}} & \rot{MaxLogit~\cite{zhang2023decoupling}} & \rot{ODIN~\cite{liang2018enhancing}} & \rot{ReAct~\cite{sun2021react}} & \rot{SHE~\cite{zhang2022out}} & \rot{Softmax~\cite{hendrycks2016baseline}} & \rot{Temperature~\cite{guo2017calibration}} & \rot{VIM~\cite{wang2022vim}} \\
	\midrule
	Gaussian & \cellcolor{blue!20}64 & \cellcolor{blue!50}65 & \cellcolor{blue!50}65 & \cellcolor{blue!50}65 & \cellcolor{blue!50}65 & \cellcolor{blue!5}37 & \cellcolor{blue!20}48 & \cellcolor{blue!50}89 & \cellcolor{blue!5}35 & \cellcolor{blue!20}48 & \cellcolor{blue!20}62 & \cellcolor{blue!50}66 & \cellcolor{blue!5}35 & \cellcolor{blue!20}50 & \cellcolor{blue!20}56 & \cellcolor{blue!5}38 & \cellcolor{blue!20}61 & \cellcolor{blue!50}65 & \cellcolor{blue!5}46 \\
	Contrast & \cellcolor{blue!20}45 & \cellcolor{blue!20}41 & \cellcolor{blue!20}41 & \cellcolor{blue!20}41 & \cellcolor{blue!20}41 & \cellcolor{blue!50}56 & \cellcolor{blue!20}44 & \cellcolor{blue!5}20 & \cellcolor{blue!50}56 & \cellcolor{blue!20}42 & \cellcolor{blue!50}64 & \cellcolor{blue!20}49 & \cellcolor{blue!50}59 & \cellcolor{blue!50}50 & \cellcolor{blue!50}51 & \cellcolor{blue!50}57 & \cellcolor{blue!20}46 & \cellcolor{blue!20}41 & \cellcolor{blue!50}50 \\
	Invert & \cellcolor{blue!20}28 & \cellcolor{blue!20}21 & \cellcolor{blue!20}21 & \cellcolor{blue!20}21 & \cellcolor{blue!20}21 & \cellcolor{blue!50}47 & \cellcolor{blue!5}0 & \cellcolor{blue!5}0 & \cellcolor{blue!20}39 & \cellcolor{blue!5}0 & \cellcolor{blue!50}100 & \cellcolor{blue!50}100 & \cellcolor{blue!50}79 & \cellcolor{blue!20}43 & \cellcolor{blue!50}92 & \cellcolor{blue!50}88 & \cellcolor{blue!50}56 & \cellcolor{blue!20}21 & \cellcolor{blue!5}0 \\
	Rotate & \cellcolor{blue!20}60 & \cellcolor{blue!50}62 & \cellcolor{blue!50}62 & \cellcolor{blue!50}61 & \cellcolor{blue!50}61 & \cellcolor{blue!5}38 & \cellcolor{blue!20}43 & \cellcolor{blue!50}79 & \cellcolor{blue!5}39 & \cellcolor{blue!20}41 & \cellcolor{blue!50}69 & \cellcolor{blue!50}67 & \cellcolor{blue!5}39 & \cellcolor{blue!20}50 & \cellcolor{blue!20}59 & \cellcolor{blue!20}41 & \cellcolor{blue!50}62 & \cellcolor{blue!50}61 & \cellcolor{blue!20}41 \\
	KMNIST & \cellcolor{blue!20}64 & \cellcolor{blue!50}82 & \cellcolor{blue!20}81 & \cellcolor{blue!20}81 & \cellcolor{blue!50}82 & \cellcolor{blue!20}18 & \cellcolor{blue!5}16 & \cellcolor{blue!50}84 & \cellcolor{blue!20}18 & \cellcolor{blue!5}10 & \cellcolor{blue!50}98 & \cellcolor{blue!50}97 & \cellcolor{blue!20}18 & \cellcolor{blue!20}54 & \cellcolor{blue!50}84 & \cellcolor{blue!20}30 & \cellcolor{blue!50}82 & \cellcolor{blue!50}82 & \cellcolor{blue!5}14 \\
	\bottomrule
\end{tabular}
	
	\label{tab:good-examples}
\end{table}

We demonstrate the necessity of having a clear evaluation in \cref{tab:good-examples}. The full table containing all available OOD datasets can be found in \cref{tb:auroc-full} in \myrefappendix{app:comparison}.
We evaluate the available monitors on a network trained on the MNIST dataset on a GPU and depict the AUROC score. 
The values of MDS and Mahalanobis can differ when switching between CPU and GPU; refer to \myrefappendix{app:cpu-gpu} for details.
The {\scshape Box} monitor~\cite{henzinger2020outside} is not included as it does not have a single threshold and, therefore, no AUROC score can be computed.
The table shows the ranking of the monitors for the detection of Gaussian noise, increased contrast, color inversion, rotation, and a new, albeit similar dataset (KMNIST). 
A darker color indicates a better ranking.
One can see that there is barely any common behavior among the monitors. 
For example, while {\scshape GradNorm} performs best on Gaussian noise,
it performs worst on inverted images. 

This also shows that it is important for the user to define her goal for the monitor. 
Not every monitor will be great at detecting a particular type of OOD, and she must carefully choose the right monitor for her setting. \monitizer eases this task.
In addition, it highlights the need for a clear evaluation of new monitoring methods in scientific publications.

We illustrate further features of \monitizer using the following four monitors:
{\scshape Energy}~\cite{liu2020energy}, {\scshape ODIN}~\cite{liang2018enhancing}, {\scshape Box}~\cite{henzinger2020outside}, and {\scshape Gaussian}~\cite{gaussMon}.
The first two were proposed by the machine-learning community, and the latter two by the formal methods community.

\begin{figure}[t]
	\centering 
	\includegraphics[width=0.8\textwidth,trim=8 8 4 0, clip]{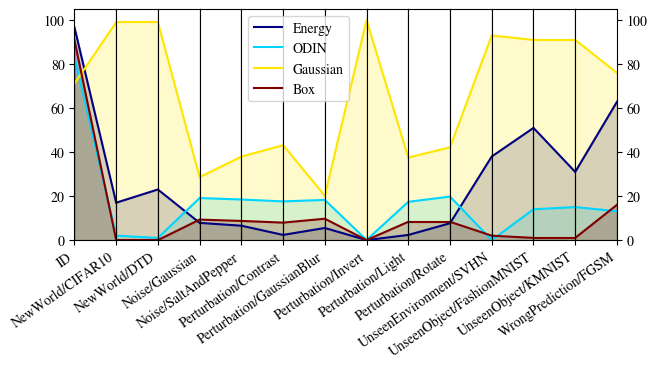}
	\caption{The monitor templates were optimized on MNIST as ID and for detecting New-World / \cifarten as OOD while keeping \SI{70}{\percent} accuracy on ID. All monitors were optimized randomly. 
	}
	\label{fig:monitors-random-cifar10}	
\end{figure}

The output produced by \monitizer in the form of tables and plots (depicted in \cref{fig:monitors-random-cifar10})
helps the user see the effect of the choice of monitor, chosen objective, and dataset
on the monitor's effectiveness.
\monitizer allows users to experiment with different choices
and select the one suitable 
for their needs.
\cref{fig:monitors-random-cifar10} shows the evaluation of the mentioned monitors with the MNIST dataset as ID data and an optimization with the goal of detecting pre-selected images of the \cifarten dataset as those are entirely unknown to the network.
The optimization was performed randomly.
This resulted in the {\scshape Gaussian} monitor only correctly classifying around 70\% of ID data, whereas the other monitors have higher accuracy on ID data. 
Consequently, the other monitors
perform worse than the {\scshape Gaussian} monitor in detecting OOD data, as there is a tradeoff between good performance on ID and OOD data.
This highlights the necessity of proper optimization for each monitor.
See \myrefappendix{app:eval} for a detailed evaluation where we report on the experiments with different monitors, optimization objectives, and datasets.

Our experiments show
that different monitors have different strengths and limitations.
One can tune a monitor for a specific purpose (e.g., detecting a particular OOD class with very high accuracy); 
however, this affects its performance in other OOD classes.

\section{Conclusion}
\label{sec:conclusion}
\monitizer is a tool for automating the design and evaluation of NN monitors.
It supports
developers of new monitoring techniques,
potential users of available monitors,
and researchers attempting to improve the state of the art.
In particular, it optimizes the monitor for the objectives specified by the user
and thoroughly evaluates it. 

\monitizer provides a library of 19 monitors, accompanied by
9 datasets and 15 NNs (at least one for each dataset),
and three optimization methods (random, grid-search, and gradient descent).
Additionally, all these inputs can be easily customized by a few lines of Python code,
allowing a user to provide their monitors, datasets, and networks.
The framework is extensible so that the user can implement
their custom optimization methods and objectives.

\monitizer is an open-source tool providing a freely available platform for new monitors
and easing their evaluation.
It is publicly available at \url{https://gitlab.com/live-lab/software/monitizer}.

\vfill

\inlineheadingbf{Data Availability Statement}
A reproduction package including all our results is available at Zenodo~\cite{Monitizer-artifact}.

\bibliographystyle{splncs04}
\bibliography{ref}

\newpage
\appendix
\section*{Appendix}
\renewcommand{\thesubsection}{\Alph{subsection}}

The structure of the appendix is as follows: 
\begin{itemize}
	\item In \cref{app:description-of-ood}, we explain the different OOD classes in more detail than in the main body.
	\item In \cref{app:optimization}, we provide additional information about the optimization step of \monitizer.
	\item In \cref{app:statistical-analysis}, we explain how one can produce confidence intervals for the computed values.
	\item \cref{app:example-use-cases} provides specific examples for each use-case to demonstrate how the respective user would use \monitizer.
	\item \cref{app:details-networks} shows the information about the networks we used for our experiments and that are provided by \monitizer.
	\item \cref{app:eval} contains a more detailed evaluation.
\end{itemize}

\subsection{Description of OOD-classes}\label{app:description-of-ood}
\begin{figure}[th!]
	\centering
	\begin{subfigure}[t]{0.15\textwidth}
		\includegraphics[width=0.9\linewidth]{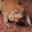}
		\caption{
		} 
		\label{fig:ood_exapmles_id}
	\end{subfigure}
	\begin{subfigure}[t]{0.15\textwidth}
		\includegraphics[width=0.9\linewidth]{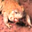}
		\caption{
		}
	\end{subfigure}
	\begin{subfigure}[t]{0.15\textwidth}
		\includegraphics[width=0.9\linewidth]{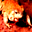}
		\caption{
		}\label{fig:ood_examples_1}
	\end{subfigure}
	\begin{subfigure}[t]{0.15\textwidth}
		\includegraphics[width=0.9\linewidth]{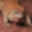}
		\caption{
		}
	\end{subfigure}\\
	\begin{subfigure}[t]{0.15\textwidth}
		\includegraphics[width=0.9\linewidth]{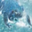}
		\caption{
		}
	\end{subfigure}
	\begin{subfigure}[t]{0.15\textwidth}
		\includegraphics[width=0.9\linewidth]{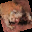}
		\caption{
		}
	\end{subfigure}
	\begin{subfigure}[t]{0.15\textwidth}
		\includegraphics[width=0.9\linewidth]{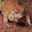}
		\caption{
		}
	\end{subfigure}
	\begin{subfigure}[t]{0.15\textwidth}
		\includegraphics[width=0.9\linewidth]{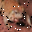}
		\caption{
		}
	\end{subfigure}\\
	\begin{subfigure}[t]{0.15\textwidth}
		\includegraphics[width=0.9\linewidth]{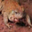}
		\caption{
		}
	\end{subfigure}
	\begin{subfigure}[t]{0.15\textwidth}
		\includegraphics[width=0.9\linewidth]{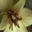}
		\caption{
		}
	\end{subfigure}
	\begin{subfigure}[t]{0.15\textwidth}
		\includegraphics[width=0.9\linewidth]{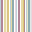}
		\caption{
		}
	\end{subfigure}	
	\begin{subfigure}[t]{0.15\textwidth}
		\includegraphics[width=0.9\linewidth]{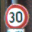}
		\caption{
		} \label{fig:ood_examples_9}
	\end{subfigure}
	\begin{subfigure}[t]{0.15\textwidth}
		\includegraphics[width=0.9\linewidth]{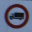}
		\caption{
		} \label{fig:ood_examples_nood_1}
	\end{subfigure}
	\caption{\textbf{Demonstration of our classification}: (a) is ID data from \cifarten~\cite{cifar}, 
		(b) - (l) are examples of different OOD classes, 
		and (m) is an image from GTSRB that should not be considered OOD.\\
		(b) Light (increased), (c) Contrast (increased), (d) Gaussian Blur, (e) Color Inversion, (f) Rotation, (g) Gaussian Noise, (h) Salt-and-Pepper Noise, (i) FGSM, (j) Unseen Object from CIFAR-100~\cite{cifar}, (k) New World from DTD~\cite{dtd}, (l) New World from GTSRB~\cite{gtsrb}	
	}
	\label{fig:ood_examples}
\end{figure}

In this section, we describe the different classes of OOD we identified. 
We provide an illustration of the classes in \cref{fig:ood_examples}.

\inlineheadingit{Perturbation}
This class contains the OOD data that is caused by some changes in the setting or environment conditions when taking an image,
e.g., as blurring, contrast, light, inversion or rotation etc..
We generate this data by employing standard image transformations.

\inlineheadingit{Noise}
In practice, the input from sensors could be corrupted by noise. 
Our current classification considers two types of noise: Gaussian Noise and Salt-and-Pepper noise.
The framework allows a user to set the parameters for introducing noise in ID data.

\inlineheadingit{Wrong Prediction}
Following the suggestions of \cite{Guerin2023}, monitors should also be able to detect wrong predictions of the NN, especially adversarial examples \cite{adversarial-examples}.
This class contains images from the ID-dataset that are changed by using adversarial attacks, in our case FGSM~\cite{FGSM}, for this purpose.
The adversarial example is computed depending on a predefined network.
Nevertheless, we are aware that there is research focusing on the detection of adversarial examples different from the usual monitoring approaches. 

\inlineheadingit{Unseen Object}
This class of distortions corresponds to objects that were not considered during the training.
This means that the network is not trained 
to detect this type of objects,
even though the new object is similar in its environment and shape to the known objects.
For example, the K-MNIST dataset~\cite{kmnist} (\cref{fig:kmnist-1}), which contains Japanese letters instead of numbers
can be considered as similar but unknown OOD data for MNIST~\cite{mnist} (\cref{fig:mnist-1}).
Another example would be to use CIFAR-100 as OOD dataset for CIFAR-10.
However, this requires to be careful to not consider overlapping classes as OOD.

\inlineheadingit{Unseen Environment}
This class of distortions corresponds to objects or type of objects already known to the network,
however, with a slight change in the environment.
For example, considering MNIST as ID data, the images from the SVHN dataset~\cite{svhn} containing house numbers taken in the real world would represent a possible OOD dataset.
The images are scaled in a similar way as known to the network,
but embedded in a different environment.
A practical example is of autonomous driving,
where the images taken during driving could be distorted due to different weather conditions.

\inlineheadingit{New World}
This class of distortions corresponds to data completely different from the data the NN being monitored was trained on.
For example, this could be an image from CIFAR-10~\cite{cifar} for a network trained on MNIST~\cite{mnist}.
This class requires special care to ensure that the OOD is actually different.
This could even mean to compare the prediction of the network with the OOD input,
as some similar information might be hidden in the OOD input.
An example illustrating this can be seen in \cref{fig:ood_examples_nood_1}.
A network trained on CIFAR-10~\cite{cifar} might still predict \textit{truck}, which is correct.
We should not consider this case when reporting the performance of the monitor on datasets of the type \textit{New World}.

\FloatBarrier

 \subsection{Optimization}\label{app:optimization}
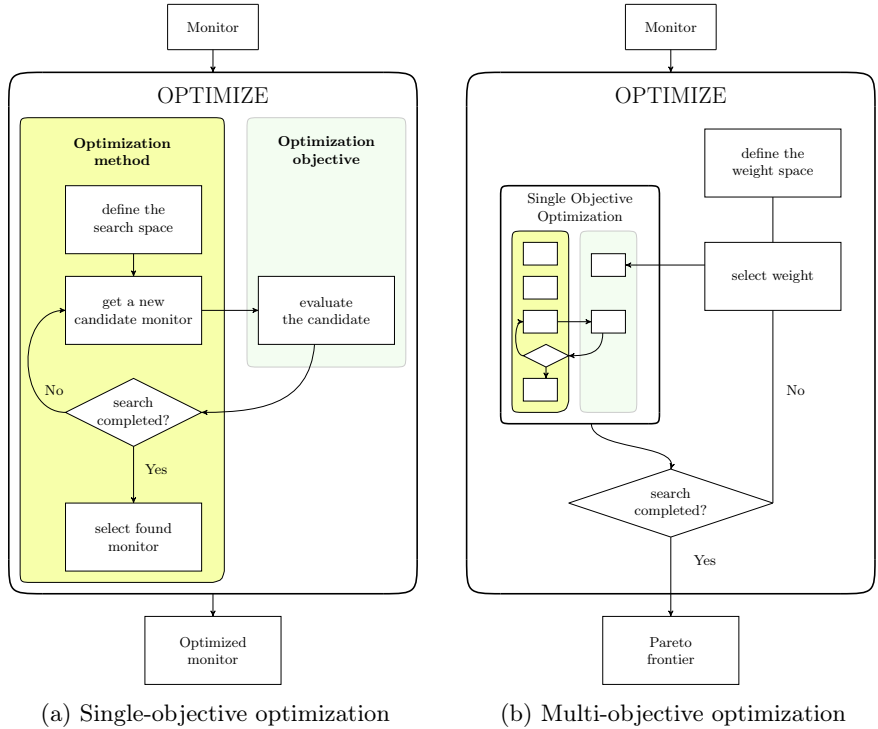
\begin{figure}[h!]
 	\begin{subfigure}{0.49\textwidth}
 		\centering
 		\resizebox{0.95\textwidth}{!}{\begin{tikzpicture}
	\begin{pgfonlayer}{nodelayer}
		\node [style=none] (103) at (-19.75, 9.75) {};
		\node [style=none] (104) at (-19.5, 10) {};
		\node [style=none] (105) at (-15.5, 10) {};
		\node [style=none] (106) at (-15.25, 9.75) {};
		\node [style=none] (107) at (-15.25, 0) {};
		\node [style=none] (108) at (-15.5, -0.25) {};
		\node [style=none] (109) at (-19.5, -0.25) {};
		\node [style=none] (110) at (-19.75, 0) {};
		\node [style=none] (0) at (-16.5, 12.5) {};
		\node [style=none] (1) at (-14.5, 12.5) {};
		\node [style=none] (2) at (-16.5, 11.5) {};
		\node [style=none] (3) at (-14.5, 11.5) {};
		\node [style=none] (4) at (-15.5, 12) {};
		\node [style=none] (5) at (-15.5, 12) {Monitor};
		\node [style=none] (62) at (-18.75, 5.75) {};
		\node [style=none] (64) at (-19, 4) {No};
		\node [style=none] (70) at (-16.75, 2.25) {Yes};
		\node [style=none] (71) at (-19.25, 11) {};
		\node [style=none] (72) at (-11.75, 11) {};
		\node [style=none] (75) at (-11, 10.25) {};
		\node [style=none] (76) at (-20, 10.25) {};
		\node [style=none] (77) at (-20, 0.25) {};
		\node [style=none] (78) at (-19.25, -0.5) {};
		\node [style=none] (79) at (-11.75, -0.5) {};
		\node [style=none] (80) at (-11, 0.25) {};
		\node [style=none] (81) at (-14.75, 9.75) {};
		\node [style=none] (82) at (-14.5, 10) {};
		\node [style=none] (83) at (-11.5, 10) {};
		\node [style=none] (84) at (-11.25, 9.75) {};
		\node [style=none] (85) at (-11.25, 4.75) {};
		\node [style=none] (86) at (-11.5, 4.5) {};
		\node [style=none] (87) at (-14.5, 4.5) {};
		\node [style=none] (88) at (-14.75, 4.75) {};
		\node [style=none] (137) at (-17.25, 4.25) {};
		\node [style=none] (138) at (-18.75, 3.5) {};
		\node [style=none] (139) at (-15.75, 3.5) {};
		\node [style=none] (140) at (-17.25, 2.75) {};
		\node [style=none] (141) at (-17.25, 1.5) {};
		\node [style=none] (142) at (-15.75, 5.75) {};
		\node [style=none] (143) at (-14.5, 5.75) {};
		\node [style=none] (144) at (-13.25, 5.25) {};
		\node [style=none] (153) at (-15.5, -0.5) {};
		\node [style=none] (156) at (-15.5, 11) {};
		\node [style=none] (157) at (-15.5, 11.5) {};
		\node [style=none] (158) at (-15.5, 10.5) {\Large OPTIMIZE};
		\node [style=center-align] (159) at (-17.5, 9.25) {\textbf{Optimization}\\\textbf{method}};
		\node [style=center-align] (160) at (-13, 9.25) {\textbf{Optimization}\\\textbf{objective}};
		\node [style=rectangle-white] (162) at (-13, 5.75) {evaluate\\the candidate};
		\node [style=rectangle-white] (163) at (-17.25, 7.75) {define the\\search space};
		\node [style=rectangle-white] (164) at (-17.25, 5.75) {get a new\\candidate monitor};
		\node [style=rectangle-white] (165) at (-17.25, 0.75) {select found\\monitor};
		\node [style=none] (166) at (-17.25, 7.5) {};
		\node [style=none] (167) at (-17.25, 6.5) {};
		\node [style=none] (168) at (-17, 7) {};
		\node [style=rectangle-white] (169) at (-15.5, -1.75) {Optimized\\monitor};
		\node [style=none] (170) at (-15.5, -1) {};
		\node [style=none] (171) at (-15.5, -0.5) {};
		\node [style=center-align] (173) at (-17.25, 3.5) {search\\completed?};
	\end{pgfonlayer}
	\begin{pgfonlayer}{edgelayer}
		\draw [style=yellow-box-line] (106.center)
			 to (107.center)
			 to [bend left, looseness=1.50] (108.center)
			 to (109.center)
			 to [bend left=45, looseness=1.75] (110.center)
			 to (103.center)
			 to [bend left=45, looseness=1.75] (104.center)
			 to (105.center)
			 to [bend left=45, looseness=1.75] cycle;
		\draw [style=line] (0.center) to (1.center);
		\draw [style=line] (1.center) to (3.center);
		\draw [style=line] (3.center) to (2.center);
		\draw [style=line] (2.center) to (0.center);
		\draw [style=thin line, bend left=45, looseness=1.75] (72.center) to (75.center);
		\draw [style=thin line] (75.center) to (80.center);
		\draw [style=thin line, bend left=45, looseness=1.75] (80.center) to (79.center);
		\draw [style=thin line] (79.center) to (78.center);
		\draw [style=thin line, bend right=315, looseness=1.75] (78.center) to (77.center);
		\draw [style=thin line] (77.center) to (76.center);
		\draw [style=thin line, bend left=45, looseness=1.75] (76.center) to (71.center);
		\draw [style=thin line] (71.center) to (72.center);
		\draw [style=green-box-line] (86.center)
			 to (87.center)
			 to [bend right=315, looseness=1.75] (88.center)
			 to (81.center)
			 to [bend left=45, looseness=1.75] (82.center)
			 to (83.center)
			 to [bend left=45, looseness=1.75] (84.center)
			 to (85.center)
			 to [bend left=45, looseness=1.75] cycle;
		\draw [style=line] (137.center)
			 to (139.center)
			 to (140.center)
			 to (138.center)
			 to cycle;
		\draw [style=arrow] (140.center) to (141.center);
		\draw [style=arrow] (142.center) to (143.center);
		\draw [style=arrow, in=0, out=-90, looseness=1.25] (144.center) to (139.center);
		\draw [style=arrow] (157.center) to (156.center);
		\draw [style=arrow 2, bend right=270, looseness=1.25] (138.center) to (62.center);
		\draw [style=arrow] (166.center) to (167.center);
		\draw [style=arrow] (171.center) to (170.center);
	\end{pgfonlayer}
\end{tikzpicture}}
 		\caption{Single-objective optimization}
 		\label{fig:single-objective-1}
 	\end{subfigure}
 	\begin{subfigure}{0.49\textwidth}
 		\centering
 		\resizebox{0.95\textwidth}{!}{\begin{tikzpicture}
	\begin{pgfonlayer}{nodelayer}
		\node [style=none] (220) at (-6.5, 12.5) {};
		\node [style=none] (221) at (-4.5, 12.5) {};
		\node [style=none] (222) at (-6.5, 11.5) {};
		\node [style=none] (223) at (-4.5, 11.5) {};
		\node [style=none] (224) at (-5.5, 12) {};
		\node [style=none] (225) at (-5.5, 12) {Monitor};
		\node [style=none] (229) at (-9.25, 11) {};
		\node [style=none] (230) at (-1.75, 11) {};
		\node [style=none] (231) at (-1, 10.25) {};
		\node [style=none] (232) at (-10, 10.25) {};
		\node [style=none] (233) at (-10, 0.25) {};
		\node [style=none] (234) at (-9.25, -0.5) {};
		\node [style=none] (235) at (-1.75, -0.5) {};
		\node [style=none] (236) at (-1, 0.25) {};
		\node [style=none] (261) at (-5.5, -0.5) {};
		\node [style=none] (262) at (-5.5, 11) {};
		\node [style=none] (263) at (-5.5, 11.5) {};
		\node [style=none] (264) at (-5.5, 10.5) {\Large OPTIMIZE};
		\node [style=rectangle-white] (274) at (-5.5, -1.75) {Pareto\\frontier};
		\node [style=none] (275) at (-5.5, -1) {};
		\node [style=none] (276) at (-5.5, 0.75) {};
		\node [style=none] (277) at (-9, 7.25) {};
		\node [style=none] (278) at (-8.75, 7.5) {};
		\node [style=none] (279) at (-8, 7.5) {};
		\node [style=none] (280) at (-7.75, 7.25) {};
		\node [style=none] (281) at (-7.75, 3.75) {};
		\node [style=none] (282) at (-8, 3.5) {};
		\node [style=none] (283) at (-8.75, 3.5) {};
		\node [style=none] (284) at (-9, 3.75) {};
		\node [style=none] (285) at (-8, 6.75) {};
		\node [style=none] (286) at (-8.75, 5.5) {};
		\node [style=none] (287) at (-9, 8.5) {};
		\node [style=none] (288) at (-6, 8.5) {};
		\node [style=none] (289) at (-5.75, 8.25) {};
		\node [style=none] (290) at (-9.25, 8.25) {};
		\node [style=none] (291) at (-9.25, 3.5) {};
		\node [style=none] (292) at (-9, 3.25) {};
		\node [style=none] (293) at (-6, 3.25) {};
		\node [style=none] (294) at (-5.75, 3.5) {};
		\node [style=none] (295) at (-7.5, 7.25) {};
		\node [style=none] (296) at (-7.25, 7.5) {};
		\node [style=none] (297) at (-6.5, 7.5) {};
		\node [style=none] (298) at (-6.25, 7.25) {};
		\node [style=none] (299) at (-6.25, 3.75) {};
		\node [style=none] (300) at (-6.5, 3.5) {};
		\node [style=none] (301) at (-7.25, 3.5) {};
		\node [style=none] (302) at (-7.5, 3.75) {};
		\node [style=none] (303) at (-7.25, 7) {};
		\node [style=none] (304) at (-6.5, 7) {};
		\node [style=none] (305) at (-7.25, 6.5) {};
		\node [style=none] (306) at (-6.5, 6.5) {};
		\node [style=none] (307) at (-7.25, 5.75) {};
		\node [style=none] (308) at (-6.5, 5.75) {};
		\node [style=none] (309) at (-7.25, 5.25) {};
		\node [style=none] (310) at (-6.5, 5.25) {};
		\node [style=none] (311) at (-8.75, 7.25) {};
		\node [style=none] (312) at (-8, 7.25) {};
		\node [style=none] (313) at (-8.75, 6.75) {};
		\node [style=none] (314) at (-8, 6.75) {};
		\node [style=none] (315) at (-8.75, 4.25) {};
		\node [style=none] (316) at (-8, 4.25) {};
		\node [style=none] (317) at (-8.75, 3.75) {};
		\node [style=none] (318) at (-8, 3.75) {};
		\node [style=none] (319) at (-8.25, 5) {};
		\node [style=none] (320) at (-8.75, 4.75) {};
		\node [style=none] (321) at (-7.75, 4.75) {};
		\node [style=none] (322) at (-8.25, 4.5) {};
		\node [style=none] (323) at (-8.25, 4.25) {};
		\node [style=none] (324) at (-8, 5.5) {};
		\node [style=none] (325) at (-7.25, 5.5) {};
		\node [style=none] (326) at (-7, 5.25) {};
		\node [style=none] (327) at (-5.5, 0.75) {};
		\node [style=none] (328) at (-8.75, 6.5) {};
		\node [style=none] (329) at (-8, 6.5) {};
		\node [style=none] (330) at (-8.75, 6) {};
		\node [style=none] (331) at (-8, 6) {};
		\node [style=none] (332) at (-8.75, 5.75) {};
		\node [style=none] (333) at (-8, 5.75) {};
		\node [style=none] (334) at (-8.75, 5.25) {};
		\node [style=none] (335) at (-8, 5.25) {};
		\node [style=none] (336) at (-5.5, 2.25) {};
		\node [style=none] (337) at (-7.75, 1.5) {};
		\node [style=none] (338) at (-3.25, 1.5) {};
		\node [style=none] (339) at (-5.5, 0.75) {};
		\node [style=none] (342) at (-3.25, 8.5) {};
		\node [style=none] (343) at (-3.25, 7) {};
		\node [style=none] (344) at (-4.5, 6.75) {};
		\node [style=none] (345) at (-6.5, 6.75) {};
		\node [style=none] (346) at (-7.25, 3.25) {};
		\node [style=none] (347) at (-3.25, 6) {};
		\node [style=none] (348) at (-4.75, 0.25) {Yes};
		\node [style=none] (349) at (-2.75, 4) {No};
		\node [style=rectangle-white] (350) at (-3.25, 9) {define the\\weight space};
		\node [style=rectangle-white] (351) at (-3.25, 6.5) {select weight};
		\node [style=center-align] (352) at (-7.5, 8) {Single Objective\\Optimization};
		\node [style=center-align] (353) at (-5.5, 1.5) {search\\completed?};
	\end{pgfonlayer}
	\begin{pgfonlayer}{edgelayer}
		\draw [style=line] (220.center) to (221.center);
		\draw [style=line] (221.center) to (223.center);
		\draw [style=line] (223.center) to (222.center);
		\draw [style=line] (222.center) to (220.center);
		\draw [style=thin line, bend left=45, looseness=1.75] (230.center) to (231.center);
		\draw [style=thin line] (231.center) to (236.center);
		\draw [style=thin line, bend left=45, looseness=1.75] (236.center) to (235.center);
		\draw [style=thin line] (235.center) to (234.center);
		\draw [style=thin line, bend right=315, looseness=1.75] (234.center) to (233.center);
		\draw [style=thin line] (233.center) to (232.center);
		\draw [style=thin line, bend left=45, looseness=1.75] (232.center) to (229.center);
		\draw [style=thin line] (229.center) to (230.center);
		\draw [style=arrow] (263.center) to (262.center);
		\draw [style=arrow] (276.center) to (275.center);
		\draw [style=yellow-box-line] (280.center)
			 to (281.center)
			 to [bend left, looseness=1.50] (282.center)
			 to (283.center)
			 to [bend left=45, looseness=1.75] (284.center)
			 to (277.center)
			 to [bend left=45, looseness=1.75] (278.center)
			 to (279.center)
			 to [bend left=45, looseness=1.75] cycle;
		\draw [style=thin line, bend left=45, looseness=1.75] (288.center) to (289.center);
		\draw [style=thin line] (289.center) to (294.center);
		\draw [style=thin line, bend left=45, looseness=1.75] (294.center) to (293.center);
		\draw [style=thin line] (293.center) to (292.center);
		\draw [style=thin line, bend right=315, looseness=1.75] (292.center) to (291.center);
		\draw [style=thin line] (291.center) to (290.center);
		\draw [style=thin line, bend left=45, looseness=1.75] (290.center) to (287.center);
		\draw [style=thin line, in=180, out=0] (287.center) to (288.center);
		\draw [style=green-box-line] (300.center)
			 to (301.center)
			 to [bend right=315, looseness=1.75] (302.center)
			 to (295.center)
			 to [bend left=45, looseness=1.75] (296.center)
			 to (297.center)
			 to [bend left=45, looseness=1.75] (298.center)
			 to (299.center)
			 to [bend left=45, looseness=1.75] cycle;
		\draw [style=line] (305.center)
			 to (303.center)
			 to (304.center)
			 to (306.center)
			 to cycle;
		\draw [style=line] (309.center)
			 to (307.center)
			 to (308.center)
			 to (310.center)
			 to cycle;
		\draw [style=line] (314.center)
			 to (313.center)
			 to (311.center)
			 to (312.center)
			 to cycle;
		\draw [style=line] (317.center)
			 to (315.center)
			 to (316.center)
			 to (318.center)
			 to cycle;
		\draw [style=line] (319.center)
			 to (321.center)
			 to (322.center)
			 to (320.center)
			 to cycle;
		\draw [style=arrow] (322.center) to (323.center);
		\draw [style=arrow] (324.center) to (325.center);
		\draw [style=arrow, in=0, out=-90, looseness=1.25] (326.center) to (321.center);
		\draw [style=arrow 2, bend left=90, looseness=0.75] (320.center) to (286.center);
		\draw [style=line] (331.center)
			 to (330.center)
			 to (328.center)
			 to (329.center)
			 to cycle;
		\draw [style=line] (335.center)
			 to (334.center)
			 to (332.center)
			 to (333.center)
			 to cycle;
		\draw [style=line] (338.center) to (339.center);
		\draw [style=line] (339.center) to (337.center);
		\draw [style=line] (337.center) to (336.center);
		\draw [style=line] (336.center) to (338.center);
		\draw [style=arrow] (342.center) to (343.center);
		\draw [style=arrow] (344.center) to (345.center);
		\draw [style=arrow, in=75, out=-90, looseness=0.75] (346.center) to (336.center);
		\draw [style=arrow] (338.center) to (347.center);
	\end{pgfonlayer}
\end{tikzpicture}}
 		\caption{Multi-objective optimization}
 		\label{fig:multi-objective}
 	\end{subfigure}
 	\caption{Optimization process in \monitizer}
 	\label{fig:optimization}
 \end{figure}

 The Optimization phase \emph{optimizes} the given monitor, i.e., tunes the parameters of a given monitor template such that an objective is maximized.
 This process is described in \cref{sec:optimization}. 
 We illustrate the principle in \cref{fig:optimization}.

 \FloatBarrier

\subsection{Confidence Intervals}
\label{app:statistical-analysis}
The evaluation of a monitor in \tool{Monitizer} is done on a set of inputs. 
Since the size of this set influences the reliability of the computed value, \tool{Monitizer} allows to compute a confidence interval:

\begin{equation}
	\hat{p}\pm Z\sqrt{\frac{\hat{p}(1-\hat{p})}{n}}
\end{equation}
where $\hat{p}$ is the prediction rate (i.e., true positive, true negative, false positive, false negative), $Z$ is the z-score of 95\%, and $n$ is the size of the evaluated dataset.

For the computation of the confidence of the AUROC, we use an \href{https://gist.github.com/doraneko94/e24643136cfb8baf03ef8a314ab9615c}{implementation} that is publicly available \cite{auroc-ci-code}.

\subsection{Examples for Use Cases}\label{app:example-use-cases}
In the main body, we mention three typical users of \monitizer (see \cref{sec:use-cases}).
In the following, we provide examples for these users.

\subsubsection{The End User}

\emph{Setting}: Let Alice be an engineer in the aviation industry, developing an autonomous drone. 
Her colleagues have trained a perception NN that detects obstacles in the air. 
They also told her that the radar, which is the most important sensor of the drone, can output noisy data if in bad weather conditions\footnote{This is an artificial example, and might not reflect the reality.}.

\emph{Goal}: Alice needs to build a monitor for this network and the radar input such that the monitor can detect noisy data. 

\emph{Required work}: Alice has to implement a loading mechanism for her dataset.
After doing so, she can simply instruct \monitizer to use the custom dataset and evaluate all existing monitors to find the best one.
Since she is not interested in the general performance of the monitor, but its ability to detect noisy data, she can use the optimization of \monitizer to return the best monitor for detecting noisy images (under the constraint that it does not label more than 80\% of the good images as noisy).

\subsubsection{The Developer of Monitors}

\emph{Setting}: Bob is a researcher at an university and he has a cool new idea for a new monitoring approach for NNs. 
He has already implemented it and tried to test it, but he would need to train many NNs and he wants a simpler solution to test and compare his approach.

\emph{Goal}: Bob wants to evaluate his monitoring approach on different datasets and compare it to other methods.

\emph{Required work}: He needs to implement the interface for his monitoring approach. 
Then, he can directly input it to \monitizer and evaluate it on one of the existing datasets and networks.

\subsubsection{The Scholar}

\emph{Setting}: Catherine is an experienced researcher who wants to explore completely new methods for monitoring. 
She wants to create a NN monitor that can detect whether a NN has correctly detected that an object has moved in a video. She already has a network that is supposed to track objects in videos.

\emph{Goal}: She wants to have a monitor that detects when the network looses track of an object.

\emph{Required work}: Catherine needs to implement her own objective function 
to define the goal of the monitor.
Additionally, she needs to change her network and the video-data in a way that it can be used only on a finite number of images in the video. 
This can be done, for example, by having the network twice.
Each copy gets a frame of the video and detects some object.

She can then use the framework of \monitizer to try out different monitors for her objective.

\subsection{Networks}\label{app:details-networks}
We present the architectures of the NN that \monitizer provides within its library.
Note that we leave out the last layer in the description, since it depends on the number of classes in the dataset.

\paragraph{MNIST}
\begin{itemize}
	\item MNIST3x100: it consists of 3 layers with 100 neurons each and the ReLU activation function
	\item MNIST-conv: it consists of two convolution layers, each followed by a max-pooling layer and the ReLU activation function, and one fully connected layer in the end. Specifically, it is Conv(channel=16, kernel=5x5)-MaxPool(2)-Conv(channel=32, kernel=5x5)-MaxPool(2)-FullyConnected(100)
\end{itemize}
\paragraph{CIFAR-10}
\begin{itemize}
	\item \cifarten-VGG11: from \cite{simonyan2014very}
	\item \cifarten-conv: It consists of three convolution layers, each combined with batch normalization and max-pooling. Specifically, it is Conv(channel=8, kernel=5)-BatchNorm(8)-ReLU-MaxPool(kernel=2, stride=2)-Conv(channel=24, kernel=3)-BatchNorm(24)-ReLU-MaxPool(kernel=2, stride=2)-Conv(channel=48, kernel=1)-BatchNorm(48)-ReLU-MaxPool(kernel=2, stride=2)-FullyConnected(120)-ReLU-FullyConnected(84)
\end{itemize}
\paragraph{GTSRB, DTD, SVHN, CIFAR-100, K-MNIST, FashionMNIST}
\begin{itemize}
	\item 
Conv(channel=100, kernel=5) - ELU(alpha=1) - MaxPool(kernel=2, stride=2) - BatchNorm(100) -  Conv(channel=150, kernel=3) - ELU(alpha=1) - MaxPool(kernel=2, stride=2) - BatchNorm(150) - Conv(channel=250, kernel=1) - ELU(alpha=1) - MaxPool(kernel=2, stride=2) - BatchNorm(250) - FullyConnected(350) - ELU(alpha=1) - BatchNorm(350)

\end{itemize}
\paragraph{ImageNet}
\monitizer contains a stored ResNet50~\cite{he2016deep}.
A user can download additional networks from the PyTorch-library. 

\FloatBarrier

\subsection{Evaluation by Case Study}
\label{app:eval}
This section contains detailed explanation of the capabilities of \monitizer using the four monitors:
Energy~\cite{liu2020energy}, {\scshape ODIN}~\cite{liang2018enhancing}, Out-of-the-{\scshape Box}~\cite{henzinger2020outside}, and {\scshape Gaussian}~\cite{gaussMon}.
The first two were proposed by the machine-learning community and the latter two by the formal methods community.

\subsubsection{Evaluation of the Selected Monitors}
We show the effectiveness of the selected monitors on the OOD classes,
and the effect of the choice of objectives on these results.

\begin{table}\centering
	\scalebox{0.8}{
		\begin{tabular}{llr}
	\toprule
	\textbf{Monitor} & \textbf{Dataset} &  \textbf{Accuracy}\\
	\midrule
	\multirow[t]{14}{*}{Box} & ID & 90.95 \\
	& NewWorld/CIFAR10 &  0.00 \\
	& NewWorld/DTD &  0.00 \\
	& Noise/Gaussian &  9.31 \\
	& Noise/SaltAndPepper &  8.71 \\
	& Perturbation/Contrast &  7.95 \\
	& Perturbation/GaussianBlur &  9.71 \\
	& Perturbation/Invert &  0.00 \\
	& Perturbation/Light &  8.32 \\
	& Perturbation/Rotate &  8.32 \\
	& UnseenEnvironment/SVHN &  2.00 \\
	& UnseenObject/FashionMNIST &  1.00 \\
	& UnseenObject/KMNIST &  1.00 \\
	& WrongPrediction/FGSM & 16.02 \\
	\cline{1-3}
	\multirow[t]{14}{*}{Energy} & ID & 96.73 \\
	& NewWorld/CIFAR10 & 17.00 \\
	& NewWorld/DTD & 23.00 \\
	& Noise/Gaussian &  7.83 \\
	& Noise/SaltAndPepper &  6.54 \\
	& Perturbation/Contrast &  2.36 \\
	& Perturbation/GaussianBlur &  5.50 \\
	& Perturbation/Invert &  0.00 \\
	& Perturbation/Light &  2.34 \\
	& Perturbation/Rotate &  7.65 \\
	& UnseenEnvironment/SVHN & 38.00 \\
	& UnseenObject/FashionMNIST & 51.00 \\
	& UnseenObject/KMNIST & 31.00 \\
	& WrongPrediction/FGSM & 62.85 \\
	\cline{1-3}
	\multirow[t]{14}{*}{Gaussian} & ID & 71.13 \\
	& NewWorld/CIFAR10 & 99.00 \\
	& NewWorld/DTD & 99.00 \\
	& Noise/Gaussian & 28.62 \\
	& Noise/SaltAndPepper & 37.94 \\
	& Perturbation/Contrast & 43.13 \\
	& Perturbation/GaussianBlur & 20.19 \\
	& Perturbation/Invert & 100.00 \\
	& Perturbation/Light & 37.50 \\
	& Perturbation/Rotate & 42.17 \\
	& UnseenEnvironment/SVHN & 93.00 \\
	& UnseenObject/FashionMNIST & 91.00 \\
	& UnseenObject/KMNIST & 91.00 \\
	& WrongPrediction/FGSM & 76.01 \\
	\cline{1-3}
	\multirow[t]{14}{*}{ODIN} & ID & 83.16 \\
	& NewWorld/CIFAR10 &  2.00 \\
	& NewWorld/DTD &  1.00 \\
	& Noise/Gaussian & 19.12 \\
	& Noise/SaltAndPepper & 18.44 \\
	& Perturbation/Contrast & 17.59 \\
	& Perturbation/GaussianBlur & 18.26 \\
	& Perturbation/Invert &  0.05 \\
	& Perturbation/Light & 17.40 \\
	& Perturbation/Rotate & 19.78 \\
	& UnseenEnvironment/SVHN &  0.00 \\
	& UnseenObject/FashionMNIST & 14.00 \\
	& UnseenObject/KMNIST & 15.00 \\
	& WrongPrediction/FGSM & 13.12 \\
	\cline{1-3}
	\bottomrule
\end{tabular}

}
	\caption{We show the accuracy (i.e. true-positive rate on OOD data and true-negative on ID data) for the four monitors, which were optimized for detecting \cifarten as OOD while keeping \SI{70}{\percent} on ID. All monitors were optimized randomly. 
	}
	\label{tab:parallel-cifar10}
\end{table}

\inlineheadingbf{Impact of the choice of monitor}
\monitizer enables a transparent evaluation of monitors on all the OOD-classes.
\cref{fig:monitors-random-cifar10} in the main part of the paper shows the performance of the monitors
for each OOD class when optimized for detecting \cifarten
as OOD while keeping \SI{70}{\percent} accuracy on the ID data.
These monitors were optimized using the random method.
\cref{tab:parallel-cifar10} shows the corresponding numbers returned by \monitizer.

The monitors have different strengths and weaknesses.
Notably, the {\scshape Gaussian} monitor excels in comparison to other monitors across all OOD classes.
Conversely, on ID data, Energy, {\scshape Box}, and {\scshape ODIN} perform better than the {\scshape Gaussian} monitor.
The {\scshape Energy} monitor shows better performance on ID than {\scshape ODIN} and the {\scshape Box} monitor.
On the OOD data, their performance varies. 
The {\scshape Box} monitor and {\scshape ODIN} perform quite similar.
Note that the {\scshape Gaussian} monitor is the only one to reliably detect inverted images in contrast to the other monitors.

\begin{figure}[t]
	\centering 
	\includegraphics[width=0.8\textwidth,trim=8 8 4 0, clip]{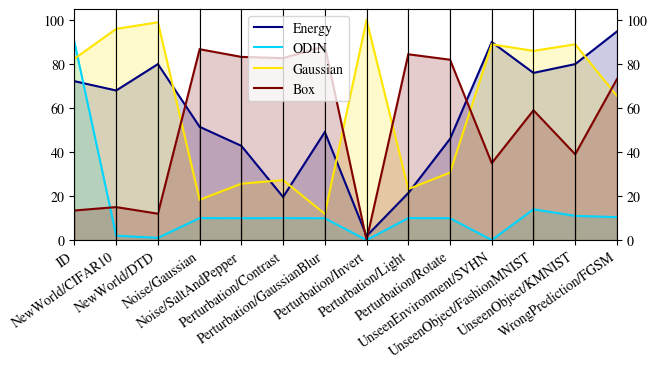}
	\caption{The monitors were optimized for detecting KMNIST as OOD while keeping a detection rate of \SI{70}{\percent} on ID, which is MNIST. {\scshape Energy} and {\scshape ODIN} are optimized with grid-search, the {\scshape Gaussian} monitor randomly. 
	}
	\label{fig:comparison-objective}
\end{figure}

\inlineheadingbf{Impact of the chosen objective}
\cref{fig:comparison-objective} shows the performance of the same set of monitors on the same OOD classes,
however, with a different optimization objective.
This time, the monitors were optimized to detect the KMNIST dataset as OOD while keeping \SI{70}{\percent} accuracy on the ID data.
{\scshape Energy} and {\scshape ODIN} were optimized using grid-search instead of random search.

The monitors exhibit variations in performance. 
Looking closer, one can see that the {\scshape Gaussian} monitor outperforms {\scshape Energy} on ID, FashionMNIST, KMNIST and on the inverted images.
However, it is noteworthy that {\scshape Energy} outperforms the {\scshape Gaussian} monitor on inputs characterized by noise ({\scshape Gaussian} and Salt-and-Pepper noise), as well as on blurry and rotated images.
{\scshape ODIN} performs better on ID but almost always worse than the other monitors on the OOD data.
The {\scshape Box} monitor performs better than the others on noisy and perturbed images, except for inverted images.

Recall the example from \cref{fig:monitor-comp-challenge}, where the {\scshape Gaussian} monitor performed worse on Gaussian Noise but better on contrast- and brightness changes.
These results were taken from \cref{fig:comparison-objective}. 
We can see, however, that the {\scshape Gaussian} monitor 
performs better on Gaussian Noise when optimized for \cifarten.
For a general comparison of OOD monitors, it its, thus, necessary to define the OOD classes for which the monitor is optimized and on which it is evaluated.

The choice of an optimization objective influences the performance of a monitor.
\monitizer allows to specify the optimization objective enabling a transparent evaluation of monitors 
for the chosen objective.
	
	\inlineheadingbf{Impact of the dataset}
	\begin{figure}[t]
		\centering
		\includegraphics[width=0.8\linewidth,trim=8 8 4 0, clip]{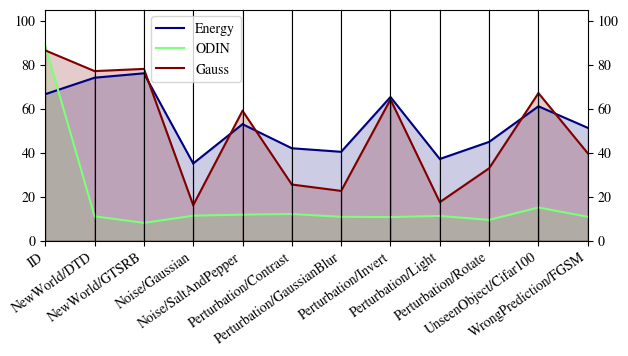}
		\caption{Optimized for detecting blurry images while keeping \SI{70}{\percent} on ID, which is \cifarten. {\scshape Energy} and {\scshape ODIN} are optimized with grid-search, the {\scshape Gaussian} monitor randomly. 
		}
		\label{fig:parallel-line-plot-on-cifar-10-ood-blur}
	\end{figure}
	We also provide experiments on \cifarten in \cref{fig:parallel-line-plot-on-cifar-10-ood-blur}.
	Unfortunately, the {\scshape Box} monitor did not work on \cifarten, since it exceeded the memory.
	We can see that in this case, the performance of {\scshape Energy} is better on an NN trained on \cifarten than on another NN trained MNIST.
	It outperforms the {\scshape Gaussian} monitor on six OOD classes, whereas the {\scshape Gaussian} monitor is only better on four classes.
	The {\scshape ODIN} monitor performs consistently worse on all OOD classes than the other two monitors, which is consistent with the case for MNIST.
	
	\inlineheadingbf{Multi-objective} 
	\begin{figure}[t]
		\centering
		\includegraphics[width=0.7\textwidth]{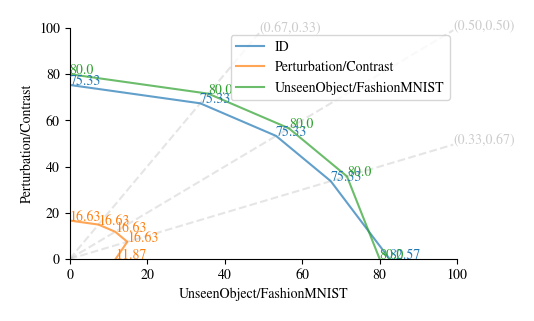}
		\caption{Pareto-curve for {\scshape Energy} optimized with grid-search (50 splits per parameter) on an NN trained on MNIST. It was optimized for CIFAR-10 and KMNIST, subject to an ID-detection of 70\%.}
	\label{fig:multi-objective-evaluation}
\end{figure}
\monitizer also supports multi-objective optimization.
\cref{fig:multi-objective-evaluation} shows an example of the Pareto-frontier for the Energy-monitor 
when optimized for two objectives with different weights. 
The labels on the gray lines describe the weights that are used for each objective. 
The corresponding value at the intersection of the colored lines with the gray lines shows the evaluated detection rate for each objective. 
This plotting of a Pareto curve follows the standard procedure of \cite{goodarzi2014introduction}.

The figure shows that the weighting of the objectives makes a difference in the limit when we only focus on CIFAR-10 as on OOD class. 
Then, we can achieve a higher detection on the ID data, but less on KMNIST.
In the other cases, the optimization procedure always finds the same balance.

\inlineheadingbf{Scalability and runtimes}
We also evaluate the runtimes of our approach.
Our experiments were executed on machines with the following configuration:
one 2.85GHz CPU (AMD EPYC 7443 24-Core Processor) with 190 GB RAM, on Ubuntu 22.04.2 LTS.
As a short summary, all our experiments run within at least \SI{105}{\second} and at most \SI{16805}{\second}.
However, this differs between datasets: on MNIST, the runtimes vary only between \SI{105}{\second} and \SI{2198}{\second}; on \cifarten, they range between \SI{142}{\second} and \SI{16805}{\second}.
Naturally, our approach scales with the size of the NN and the optimization objective.
The bigger the network, the longer one forward-pass, the longer the optimization.
Similarly, the more complicated it is to compute the objective, the longer the whole optimization.
To conclude, we cannot guarantee any runtime, since it completely depends on the inputs and monitors.

\FloatBarrier
\subsection{Performance of All Implemented Monitors}\label{app:comparison}
In this section, we present the performance of all implemented monitors on all OOD classes, which extends \cref{tab:good-examples} in \cref{tb:auroc-full}.
\NewDocumentCommand{\rot}{O{90} O{1em} m}{\makebox[#2][l]{\rotatebox{#1}{#3}}}%
\begin{table}
	\caption{Comparison of the AUROC-score of all implemented monitors. They were evaluated on a fully-connected network trained on MNIST. We round to two digits after decimal and show percentage.}
	\label{tb:auroc-full}
	\centering
	\begin{tabular}{crrrrrrrrrrrrrrrrrrr}
		\toprule
		Perturbations & \rot{ASH-B~\cite{djurisic2022extremely}} & \rot{ASH-P~\cite{djurisic2022extremely}} & \rot{ASH-S~\cite{djurisic2022extremely}} & \rot{DICE~\cite{sun2022dice}} & \rot{Energy~\cite{liu2020energy}} & \rot{Entropy~\cite{macedo2021entropic}} & \rot{Gauss~\cite{gaussMon}} & \rot{GradNorm~\cite{huang2021importance}} & \rot{KL Matching~\cite{hendrycks2022scaling}} & \rot{KNN~\cite{sun2022out}} & \rot{MDS~\cite{lee2018simple}} & \rot{Mahalanobis~\cite{ren2021simple}} & \rot{MaxLogit~\cite{zhang2023decoupling}} & \rot{ODIN~\cite{liang2018enhancing}} & \rot{ReAct~\cite{sun2021react}} & \rot{SHE~\cite{zhang2022out}} & \rot{Softmax~\cite{hendrycks2016baseline}} & \rot{Temperature~\cite{guo2017calibration}} & \rot{VIM~\cite{wang2022vim}} \\
		\midrule
		Gaussian & \cellcolor{blue!20}64 & \cellcolor{blue!50}65 & \cellcolor{blue!50}65 & \cellcolor{blue!50}65 & \cellcolor{blue!50}65 & \cellcolor{blue!5}37 & \cellcolor{blue!20}48 & \cellcolor{blue!50}89 & \cellcolor{blue!5}35 & \cellcolor{blue!20}48 & \cellcolor{blue!20}62 & \cellcolor{blue!50}66 & \cellcolor{blue!5}35 & \cellcolor{blue!20}50 & \cellcolor{blue!20}56 & \cellcolor{blue!5}38 & \cellcolor{blue!20}61 & \cellcolor{blue!50}65 & \cellcolor{blue!5}46 \\
		SaltAndPepper & \cellcolor{blue!20}57 & \cellcolor{blue!50}59 & \cellcolor{blue!50}59 & \cellcolor{blue!50}59 & \cellcolor{blue!50}59 & \cellcolor{blue!5}41 & \cellcolor{blue!20}45 & \cellcolor{blue!50}76 & \cellcolor{blue!5}41 & \cellcolor{blue!5}42 & \cellcolor{blue!50}75 & \cellcolor{blue!50}70 & \cellcolor{blue!5}41 & \cellcolor{blue!20}50 & \cellcolor{blue!20}56 & \cellcolor{blue!5}43 & \cellcolor{blue!20}57 & \cellcolor{blue!50}59 & \cellcolor{blue!20}44 \\
		Contrast & \cellcolor{blue!20}45 & \cellcolor{blue!20}41 & \cellcolor{blue!20}41 & \cellcolor{blue!20}41 & \cellcolor{blue!20}41 & \cellcolor{blue!50}56 & \cellcolor{blue!20}44 & \cellcolor{blue!5}20 & \cellcolor{blue!50}56 & \cellcolor{blue!20}42 & \cellcolor{blue!50}64 & \cellcolor{blue!20}49 & \cellcolor{blue!50}59 & \cellcolor{blue!50}50 & \cellcolor{blue!50}51 & \cellcolor{blue!50}57 & \cellcolor{blue!20}46 & \cellcolor{blue!20}41 & \cellcolor{blue!50}50 \\
		GaussianBlur & \cellcolor{blue!50}61 & \cellcolor{blue!50}64 & \cellcolor{blue!50}64 & \cellcolor{blue!50}64 & \cellcolor{blue!50}64 & \cellcolor{blue!5}38 & \cellcolor{blue!20}53 & \cellcolor{blue!50}88 & \cellcolor{blue!5}36 & \cellcolor{blue!20}55 & \cellcolor{blue!5}28 & \cellcolor{blue!20}54 & \cellcolor{blue!5}36 & \cellcolor{blue!20}50 & \cellcolor{blue!20}49 & \cellcolor{blue!5}38 & \cellcolor{blue!20}60 & \cellcolor{blue!50}64 & \cellcolor{blue!20}49 \\
		Invert & \cellcolor{blue!20}28 & \cellcolor{blue!20}21 & \cellcolor{blue!20}21 & \cellcolor{blue!20}21 & \cellcolor{blue!20}21 & \cellcolor{blue!50}47 & \cellcolor{blue!5}0 & \cellcolor{blue!5}0 & \cellcolor{blue!20}39 & \cellcolor{blue!5}0 & \cellcolor{blue!50}100 & \cellcolor{blue!50}100 & \cellcolor{blue!50}79 & \cellcolor{blue!20}43 & \cellcolor{blue!50}92 & \cellcolor{blue!50}88 & \cellcolor{blue!50}56 & \cellcolor{blue!20}21 & \cellcolor{blue!5}0 \\
		Rotate & \cellcolor{blue!20}60 & \cellcolor{blue!50}62 & \cellcolor{blue!50}62 & \cellcolor{blue!50}61 & \cellcolor{blue!50}61 & \cellcolor{blue!5}38 & \cellcolor{blue!20}43 & \cellcolor{blue!50}79 & \cellcolor{blue!5}39 & \cellcolor{blue!20}41 & \cellcolor{blue!50}69 & \cellcolor{blue!50}67 & \cellcolor{blue!5}39 & \cellcolor{blue!20}50 & \cellcolor{blue!20}59 & \cellcolor{blue!20}41 & \cellcolor{blue!50}62 & \cellcolor{blue!50}61 & \cellcolor{blue!20}41 \\
		Light & \cellcolor{blue!20}47 & \cellcolor{blue!20}43 & \cellcolor{blue!20}43 & \cellcolor{blue!20}43 & \cellcolor{blue!20}43 & \cellcolor{blue!50}55 & \cellcolor{blue!20}46 & \cellcolor{blue!5}27 & \cellcolor{blue!50}55 & \cellcolor{blue!20}45 & \cellcolor{blue!50}57 & \cellcolor{blue!20}48 & \cellcolor{blue!50}56 & \cellcolor{blue!50}50 & \cellcolor{blue!50}51 & \cellcolor{blue!50}55 & \cellcolor{blue!20}47 & \cellcolor{blue!20}43 & \cellcolor{blue!50}50 \\
		CIFAR10 & \cellcolor{blue!20}57 & \cellcolor{blue!50}75 & \cellcolor{blue!50}75 & \cellcolor{blue!50}75 & \cellcolor{blue!50}75 & \cellcolor{blue!5}22 & \cellcolor{blue!5}6 & \cellcolor{blue!5}0 & \cellcolor{blue!20}31 & \cellcolor{blue!5}2 & \cellcolor{blue!50}100 & \cellcolor{blue!50}100 & \cellcolor{blue!20}25 & \cellcolor{blue!20}44 & \cellcolor{blue!50}86 & \cellcolor{blue!20}35 & \cellcolor{blue!50}78 & \cellcolor{blue!50}75 & \cellcolor{blue!5}3 \\
		SVHN & \cellcolor{blue!20}82 & \cellcolor{blue!50}91 & \cellcolor{blue!50}90 & \cellcolor{blue!50}90 & \cellcolor{blue!50}91 & \cellcolor{blue!5}10 & \cellcolor{blue!5}6 & \cellcolor{blue!20}63 & \cellcolor{blue!20}14 & \cellcolor{blue!5}9 & \cellcolor{blue!50}98 & \cellcolor{blue!50}98 & \cellcolor{blue!5}9 & \cellcolor{blue!20}45 & \cellcolor{blue!50}92 & \cellcolor{blue!20}18 & \cellcolor{blue!20}89 & \cellcolor{blue!50}91 & \cellcolor{blue!5}10 \\
		FashionMNIST & \cellcolor{blue!20}72 & \cellcolor{blue!50}78 & \cellcolor{blue!50}78 & \cellcolor{blue!20}77 & \cellcolor{blue!50}78 & \cellcolor{blue!5}23 & \cellcolor{blue!5}14 & \cellcolor{blue!20}60 & \cellcolor{blue!5}16 & \cellcolor{blue!5}10 & \cellcolor{blue!50}97 & \cellcolor{blue!50}90 & \cellcolor{blue!5}22 & \cellcolor{blue!20}62 & \cellcolor{blue!50}80 & \cellcolor{blue!20}28 & \cellcolor{blue!50}81 & \cellcolor{blue!50}78 & \cellcolor{blue!20}27 \\
		KMNIST & \cellcolor{blue!20}64 & \cellcolor{blue!50}82 & \cellcolor{blue!20}81 & \cellcolor{blue!20}81 & \cellcolor{blue!50}82 & \cellcolor{blue!20}18 & \cellcolor{blue!5}16 & \cellcolor{blue!50}84 & \cellcolor{blue!20}18 & \cellcolor{blue!5}10 & \cellcolor{blue!50}98 & \cellcolor{blue!50}97 & \cellcolor{blue!20}18 & \cellcolor{blue!20}54 & \cellcolor{blue!50}84 & \cellcolor{blue!20}30 & \cellcolor{blue!50}82 & \cellcolor{blue!50}82 & \cellcolor{blue!5}14 \\
		FGSM & \cellcolor{blue!20}87 & \cellcolor{blue!50}94 & \cellcolor{blue!50}94 & \cellcolor{blue!50}94 & \cellcolor{blue!50}94 & \cellcolor{blue!5}7 & \cellcolor{blue!5}21 & \cellcolor{blue!50}100 & \cellcolor{blue!5}6 & \cellcolor{blue!20}27 & \cellcolor{blue!20}82 & \cellcolor{blue!20}88 & \cellcolor{blue!5}6 & \cellcolor{blue!20}48 & \cellcolor{blue!20}87 & \cellcolor{blue!5}9 & \cellcolor{blue!50}92 & \cellcolor{blue!50}94 & \cellcolor{blue!20}23 \\
		\bottomrule
	\end{tabular}
\end{table}

\subsection{MDS and Mahalanobis on CPU and GPU}\label{app:cpu-gpu}
Our experiments revealed that the floating point numbers on CPUs and GPUs can differ up to 1e-6. 
While this is mostly not an issue, it is for one standard function of the python-package ``scikit-learn'', i.e. the function ``empirical\_covariance''.
This computes an empirical covariance matrix for given inputs and can have very different outputs for a small change in the input (for reference, up to a total difference of 1e4).
Therefore, \cref{tab:good-examples} and \cref{tb:auroc-full} will look different when run on a machine with CPU.
We provide the alternative table run on CPU in \cref{tab:auroc-full-cpu} .

\begin{table}
	\caption{Comparison of the AUROC-score of all implemented monitors. They were evaluated on a fully-connected network trained on MNIST. We round to two digits after decimal and show percentage. The monitors were evaluated on CPU (makes a difference for MDS and Mahalanobis).}
	\label{tab:auroc-full-cpu}
	\centering
	\begin{tabular}{crrrrrrrrrrrrrrrrrrr}
		\toprule
		Perturbations & \rot{ASH-B~\cite{djurisic2022extremely}} & \rot{ASH-P~\cite{djurisic2022extremely}} & \rot{ASH-S~\cite{djurisic2022extremely}} & \rot{DICE~\cite{sun2022dice}} & \rot{Energy~\cite{liu2020energy}} & \rot{Entropy~\cite{macedo2021entropic}} & \rot{Gauss~\cite{gaussMon}} & \rot{GradNorm~\cite{huang2021importance}} & \rot{KL Matching~\cite{hendrycks2022scaling}} & \rot{KNN~\cite{sun2022out}} & \rot{MDS~\cite{lee2018simple}} & \rot{Mahalanobis~\cite{ren2021simple}} & \rot{MaxLogit~\cite{zhang2023decoupling}} & \rot{ODIN~\cite{liang2018enhancing}} & \rot{ReAct~\cite{sun2021react}} & \rot{SHE~\cite{zhang2022out}} & \rot{Softmax~\cite{hendrycks2016baseline}} & \rot{Temperature~\cite{guo2017calibration}} & \rot{VIM~\cite{wang2022vim}} \\
		\midrule
		Gaussian & \cellcolor{blue!50}64 & \cellcolor{blue!50}65 & \cellcolor{blue!50}65 & \cellcolor{blue!50}65 & \cellcolor{blue!50}65 & \cellcolor{blue!5}37 & \cellcolor{blue!20}48 & \cellcolor{blue!50}89 & \cellcolor{blue!5}35 & \cellcolor{blue!20}48 & \cellcolor{blue!20}55 & \cellcolor{blue!20}55 & \cellcolor{blue!5}35 & \cellcolor{blue!20}50 & \cellcolor{blue!20}56 & \cellcolor{blue!5}38 & \cellcolor{blue!20}61 & \cellcolor{blue!50}65 & \cellcolor{blue!5}46 \\
		SaltAndPepper & \cellcolor{blue!50}57 & \cellcolor{blue!50}59 & \cellcolor{blue!50}59 & \cellcolor{blue!50}59 & \cellcolor{blue!50}59 & \cellcolor{blue!5}41 & \cellcolor{blue!20}45 & \cellcolor{blue!50}76 & \cellcolor{blue!5}41 & \cellcolor{blue!5}42 & \cellcolor{blue!20}51 & \cellcolor{blue!20}52 & \cellcolor{blue!5}41 & \cellcolor{blue!20}50 & \cellcolor{blue!20}56 & \cellcolor{blue!5}43 & \cellcolor{blue!50}57 & \cellcolor{blue!50}59 & \cellcolor{blue!20}44 \\
		Contrast & \cellcolor{blue!20}45 & \cellcolor{blue!20}41 & \cellcolor{blue!20}41 & \cellcolor{blue!20}41 & \cellcolor{blue!20}41 & \cellcolor{blue!50}56 & \cellcolor{blue!20}44 & \cellcolor{blue!5}20 & \cellcolor{blue!50}56 & \cellcolor{blue!20}42 & \cellcolor{blue!20}46 & \cellcolor{blue!20}46 & \cellcolor{blue!50}59 & \cellcolor{blue!50}50 & \cellcolor{blue!50}51 & \cellcolor{blue!50}57 & \cellcolor{blue!20}46 & \cellcolor{blue!20}41 & \cellcolor{blue!50}50 \\
		GaussianBlur & \cellcolor{blue!50}61 & \cellcolor{blue!50}64 & \cellcolor{blue!50}64 & \cellcolor{blue!50}64 & \cellcolor{blue!50}64 & \cellcolor{blue!5}38 & \cellcolor{blue!20}53 & \cellcolor{blue!50}88 & \cellcolor{blue!5}36 & \cellcolor{blue!20}55 & \cellcolor{blue!20}54 & \cellcolor{blue!20}55 & \cellcolor{blue!5}36 & \cellcolor{blue!20}50 & \cellcolor{blue!20}49 & \cellcolor{blue!5}38 & \cellcolor{blue!20}60 & \cellcolor{blue!50}64 & \cellcolor{blue!20}49 \\
		Invert & \cellcolor{blue!20}28 & \cellcolor{blue!20}21 & \cellcolor{blue!20}21 & \cellcolor{blue!20}21 & \cellcolor{blue!20}21 & \cellcolor{blue!50}47 & \cellcolor{blue!5}0 & \cellcolor{blue!5}0 & \cellcolor{blue!50}39 & \cellcolor{blue!5}0 & \cellcolor{blue!20}12 & \cellcolor{blue!20}12 & \cellcolor{blue!50}79 & \cellcolor{blue!50}43 & \cellcolor{blue!50}92 & \cellcolor{blue!50}88 & \cellcolor{blue!50}56 & \cellcolor{blue!20}21 & \cellcolor{blue!5}0 \\
		Rotate & \cellcolor{blue!20}60 & \cellcolor{blue!50}62 & \cellcolor{blue!50}62 & \cellcolor{blue!50}61 & \cellcolor{blue!50}61 & \cellcolor{blue!5}38 & \cellcolor{blue!20}43 & \cellcolor{blue!50}79 & \cellcolor{blue!5}39 & \cellcolor{blue!20}41 & \cellcolor{blue!20}52 & \cellcolor{blue!20}52 & \cellcolor{blue!5}39 & \cellcolor{blue!20}50 & \cellcolor{blue!20}59 & \cellcolor{blue!20}41 & \cellcolor{blue!50}62 & \cellcolor{blue!50}61 & \cellcolor{blue!20}41 \\
		Light & \cellcolor{blue!20}47 & \cellcolor{blue!20}43 & \cellcolor{blue!20}43 & \cellcolor{blue!20}43 & \cellcolor{blue!20}43 & \cellcolor{blue!50}55 & \cellcolor{blue!20}46 & \cellcolor{blue!5}27 & \cellcolor{blue!50}55 & \cellcolor{blue!20}45 & \cellcolor{blue!20}47 & \cellcolor{blue!20}47 & \cellcolor{blue!50}56 & \cellcolor{blue!50}50 & \cellcolor{blue!50}51 & \cellcolor{blue!50}55 & \cellcolor{blue!20}47 & \cellcolor{blue!20}43 & \cellcolor{blue!50}50 \\
		CIFAR10 & \cellcolor{blue!20}57 & \cellcolor{blue!50}75 & \cellcolor{blue!50}75 & \cellcolor{blue!50}75 & \cellcolor{blue!50}75 & \cellcolor{blue!5}22 & \cellcolor{blue!5}6 & \cellcolor{blue!5}0 & \cellcolor{blue!20}31 & \cellcolor{blue!5}2 & \cellcolor{blue!20}46 & \cellcolor{blue!20}47 & \cellcolor{blue!20}25 & \cellcolor{blue!20}44 & \cellcolor{blue!50}86 & \cellcolor{blue!20}35 & \cellcolor{blue!50}78 & \cellcolor{blue!50}75 & \cellcolor{blue!5}3 \\
		SVHN & \cellcolor{blue!20}82 & \cellcolor{blue!50}91 & \cellcolor{blue!50}90 & \cellcolor{blue!50}90 & \cellcolor{blue!50}91 & \cellcolor{blue!5}10 & \cellcolor{blue!5}6 & \cellcolor{blue!20}63 & \cellcolor{blue!20}14 & \cellcolor{blue!5}9 & \cellcolor{blue!20}55 & \cellcolor{blue!20}54 & \cellcolor{blue!5}9 & \cellcolor{blue!20}45 & \cellcolor{blue!50}92 & \cellcolor{blue!20}18 & \cellcolor{blue!50}89 & \cellcolor{blue!50}91 & \cellcolor{blue!5}10 \\
		FashionMNIST & \cellcolor{blue!20}72 & \cellcolor{blue!50}78 & \cellcolor{blue!50}78 & \cellcolor{blue!50}77 & \cellcolor{blue!50}78 & \cellcolor{blue!5}23 & \cellcolor{blue!5}14 & \cellcolor{blue!20}60 & \cellcolor{blue!5}16 & \cellcolor{blue!5}10 & \cellcolor{blue!20}54 & \cellcolor{blue!20}48 & \cellcolor{blue!5}22 & \cellcolor{blue!20}62 & \cellcolor{blue!50}80 & \cellcolor{blue!20}28 & \cellcolor{blue!50}81 & \cellcolor{blue!50}78 & \cellcolor{blue!20}27 \\
		KMNIST & \cellcolor{blue!20}64 & \cellcolor{blue!50}82 & \cellcolor{blue!50}81 & \cellcolor{blue!50}81 & \cellcolor{blue!50}82 & \cellcolor{blue!20}18 & \cellcolor{blue!5}16 & \cellcolor{blue!50}84 & \cellcolor{blue!20}18 & \cellcolor{blue!5}10 & \cellcolor{blue!20}53 & \cellcolor{blue!20}50 & \cellcolor{blue!20}18 & \cellcolor{blue!20}54 & \cellcolor{blue!50}84 & \cellcolor{blue!20}30 & \cellcolor{blue!50}82 & \cellcolor{blue!50}82 & \cellcolor{blue!5}14 \\
		FGSM & \cellcolor{blue!20}87 & \cellcolor{blue!50}94 & \cellcolor{blue!50}94 & \cellcolor{blue!50}94 & \cellcolor{blue!50}94 & \cellcolor{blue!5}7 & \cellcolor{blue!5}21 & \cellcolor{blue!50}100 & \cellcolor{blue!5}6 & \cellcolor{blue!20}27 & \cellcolor{blue!20}61 & \cellcolor{blue!20}62 & \cellcolor{blue!5}6 & \cellcolor{blue!20}48 & \cellcolor{blue!20}87 & \cellcolor{blue!5}9 & \cellcolor{blue!50}92 & \cellcolor{blue!50}94 & \cellcolor{blue!20}23 \\
		\bottomrule
	\end{tabular}
\end{table}

\end{document}